\documentclass[10pt,twocolumn,letterpaper]{article}

\usepackage[pagenumbers]{cvpr} 

\usepackage{times}
\usepackage{epsfig}
\usepackage{graphicx}
\usepackage{amsmath}
\usepackage{amssymb}
\usepackage{lipsum}  
\usepackage{tabularx}
\usepackage{makecell}
\usepackage{multirow}
\usepackage{booktabs}
\usepackage{tabularx}
\usepackage{caption,soul}
\usepackage{capt-of,etoolbox}
\usepackage[pagebackref,breaklinks,colorlinks]{hyperref}
\usepackage[dvipsnames]{xcolor}
\usepackage{float}
\usepackage{enumitem}
\usepackage[accsupp]{axessibility}  

\newcommand\norm[1]{\left\lVert#1\right\rVert}
\setlist[itemize]{noitemsep}

\begin{document}

\title{ST-MFNet: A Spatio-Temporal Multi-Flow Network for Frame Interpolation}

\author{Duolikun Danier \qquad\qquad Fan Zhang \qquad\qquad David Bull\\
University of Bristol\\
{\tt\small \{duolikun.danier, fan.zhang, dave.bull\}@bristol.ac.uk}
}
\maketitle

\begin{abstract}
   Video frame interpolation (VFI) is currently a very active research topic, with applications spanning computer vision, post production and video encoding. VFI can be extremely challenging, particularly in sequences containing large motions, occlusions or dynamic textures, where existing approaches fail to offer perceptually robust interpolation performance. In this context, we present a novel deep learning based VFI method, ST-MFNet, based on a Spatio-Temporal Multi-Flow architecture. ST-MFNet employs a new multi-scale multi-flow predictor to estimate many-to-one intermediate flows, which are combined with conventional one-to-one optical flows to capture both large and complex motions. In order to enhance interpolation performance for various textures, a 3D CNN is also employed to model the content dynamics over an extended temporal window. Moreover, ST-MFNet has been trained within an ST-GAN framework, which was originally developed for texture synthesis, with the aim of further improving perceptual interpolation quality. Our approach has been comprehensively evaluated -- compared with fourteen state-of-the-art VFI algorithms -- clearly demonstrating that ST-MFNet consistently outperforms these benchmarks on varied and representative test datasets, with significant gains up to 1.09dB in PSNR for cases including large motions and dynamic textures. Our source code is available at \url{https://github.com/danielism97/ST-MFNet}.
\end{abstract}

\section{Introduction}


Video frame interpolation (VFI) has been extensively employed to deliver an improved user experience across a wide range of important applications. VFI increases the temporal resolution (frame rate) of a video through synthesizing intermediate frames between every two consecutive original frames. It can mitigate the need for costly high frame rate acquisition processes~\cite{kalluri2020flavr}, enhance the rendering of slow-motion content~\cite{jiang2018super}, support view synthesis~\cite{flynn2016deepstereo} and improve rate-quality trade-offs in video coding~\cite{wu2018video}.

In recent years, deep learning has empowered a variety of VFI algorithms. These methods can be categorized as flow-based~\cite{jiang2018super, xu2019quadratic} or kernel-based~\cite{niklaus2017video, lee2020adacof}. While flow-based methods use the estimated optical flow maps to warp input frames, kernel-based methods learn local or shared convolution kernels for synthesizing the output. To handle challenging scenarios encountered in VFI applications, various techniques have been employed to enhance these methods, including non-linear motion models~\cite{xu2019quadratic, sim2021xvfi, park2021asymmetric}, coarse-to-fine architectures~\cite{park2020bmbc, sim2021xvfi, chen2021pdwn, zhang2020flexible}, attention mechanisms~\cite{choi2020channel, kalluri2020flavr}, and deformable convolutions~\cite{lee2020adacof, gui2020featureflow}. 
\begin{figure}[t]
\begin{center}
   \includegraphics[width=\linewidth]{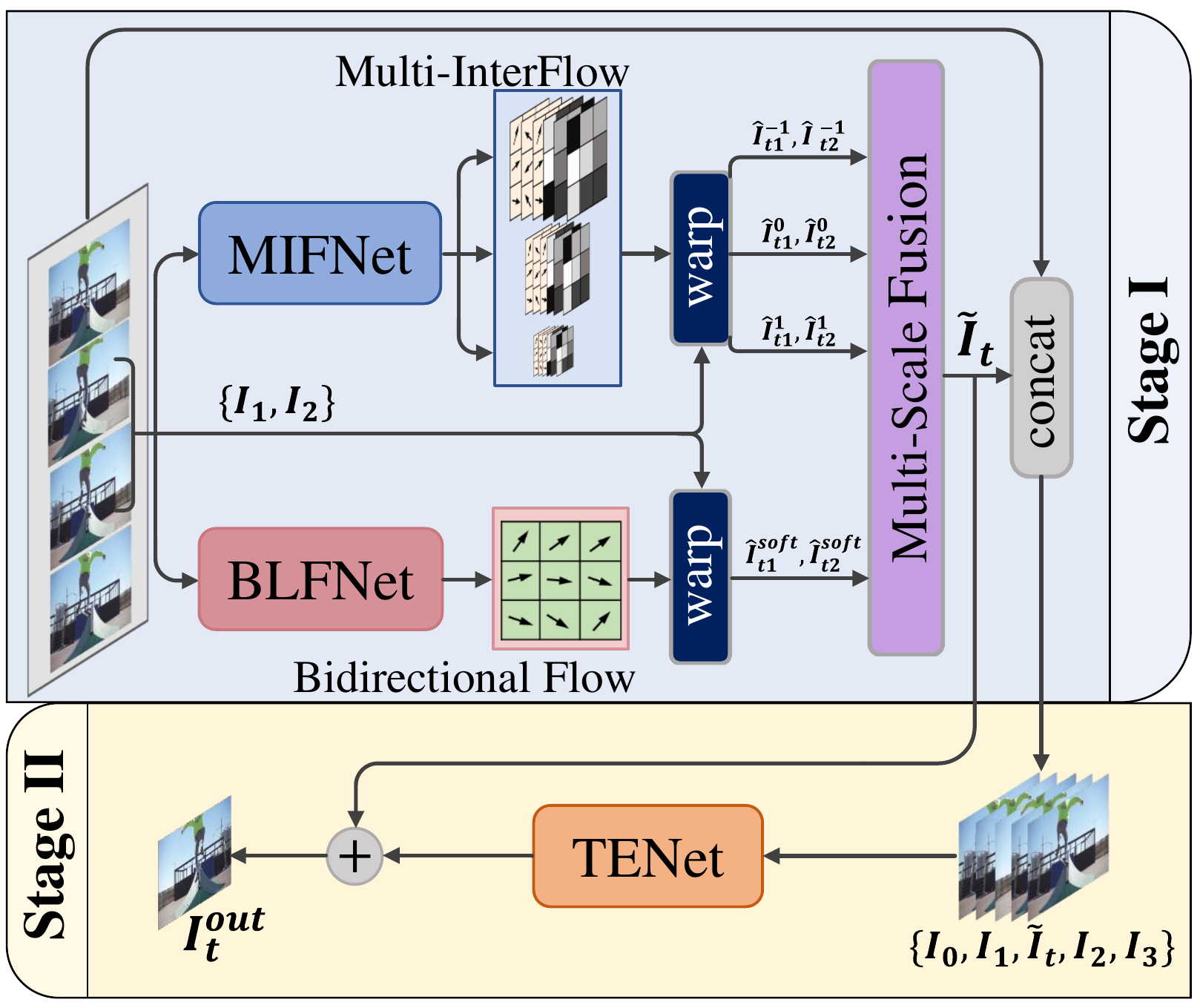}
\end{center}
\vspace{-6mm}
\caption{\label{fig:overview}
High-level architecture of ST-MFNet, which employs a two-stage workflow to interpolate an intermediate frame.}
\vspace{-5mm}
\end{figure}

Although these methods have significantly improved performance compared with conventional VFI approaches~\cite{baker2011database}, their performance can still be inconsistent, especially for content exhibiting large motions, occlusions and dynamic textures. Large motion typically means large pixel displacements, which are difficult to capture using Convolutional Neural Networks (CNNs) with limited receptive fields~\cite{adaconv, niklaus2017video}. In the case of occlusion, pixels relating to occluded objects will not appear in all input frames, thus preventing interpolation algorithms from accurately estimating the intermediate locations of those pixels~\cite{choi2020channel,kalluri2020flavr}. Finally, \textit{dynamic textures} (e.g. water, fire, foliage, etc.) exhibit more complex motion characteristics compared to the movements of rigid objects~\cite{zhang2011parametric,tafi1}. Typically, they are spatially irregular and temporally stochastic, causing most existing VFI methods to fail, especially those based on optical flow\cite{liu2017video, jiang2018super}. 

To solve these problems, we propose a novel video frame interpolation model, the Spatio-Temporal Multi-Flow Network (ST-MFNet), which decouples the handling of large and complex motions using single- and multi-flows respectively in a multi-branch
structure to offer improved interpolation performance across a wide range of content types. Specifically, ST-MFNet employs a two-stage architecture, as shown in Figure \ref{fig:overview}. In Stage I, the Multi-InterFlow Network (MIFNet) first predicts multi-interflows~\cite{dai2017deformable,lee2020adacof} at multiple scales (including an up-sampling scale simulating sub-pixel motion estimation), using a customized CNN architecture, UMSResNext, with variable kernel sizes. The multi-flows here correspond to a many-to-one mapping which enables more flexible transformation, facilitating the modeling of complex motions. To further improve the performance for large motions, a Bi-directional Linear Flow Network (BLFNet) is employed to linearly approximate the intermediate flows based on the bi-directional flows between input frames, which are estimated using a coarse-to-fine architecture~\cite{sun2018pwc}. In the second stage, inspired by recent work on texture synthesis~\cite{xie2019learning, yang2021spatiotemporal}, we integrate a 3D CNN, Texture Enhancement Network (TENet) that performs spatial and temporal filtering to capture longer-range dynamics and to predict textural residuals. Finally, we trained our model based on the ST-GAN~\cite{yang2021spatiotemporal} methodology, which was originally proposed for texture synthesis. This ensures both spatial consistency and temporal coherence of interpolated content. Extensive quantitative and qualitative studies have been performed which demonstrate the superior performance of ST-MFNet over current state-of-the-art VFI methods on a wide range of test data including large and complex motions and dynamic textures.

The primary contributions of this work are:
\begin{itemize}[noitemsep,nolistsep,leftmargin=*]
    \item A novel VFI method where multi-flow based (MIFNet) and single-flow based warping (BLFNet) are combined to enhance the capturing of complex and large motions.
    \item A new CNN architecture (UMSResNext) for the MIFNet, which predicts multiple intermediate flows at various scales, including an up-sampling scale for high precision sub-pixel motion estimation.
    \item The use of a spatio-temporal CNN (TENet) and ST-GAN, which were originally designed for texture synthesis, to enhance the interpolation of complex textures.
    \item Validation, through comprehensive experiments, that our model consistently outperforms state-of-the-art VFI methods on various scenarios, including large and complex motions and various texture types.
\end{itemize}

\section{Related Work}

In this section, we summarize recent advances in video frame interpolation (VFI) and then briefly introduce examples of dynamic texture synthesis, which have inspired the development of our method.

\subsection{Video Frame Interpolation}

Most existing VFI methods can be classified as: 

\noindent\textbf{Flow-based VFI.} This class typically involves two steps: optical flow estimation and image warping. Input frames, $I_1$ and $I_2$, are warped to a target temporal location $t$ based on either the intermediate optical flows $F_{t\rightarrow 1}, F_{t\rightarrow 2}$ (backward warping~\cite{jaderberg2015spatial}), or $F_{1\rightarrow t}, F_{2\rightarrow t}$ (forward warping~\cite{niklaus2018context}). These flows can be approximated from bi-directional optical flows ($F_{1\rightarrow 2}$ and $F_{2\rightarrow 1}$) between the input frames~\cite{jiang2018super, reda2019unsupervised, bao2019memc, bao2019depth, xu2019quadratic, niklaus2018context, niklaus2020softmax, liu2020enhanced, sim2021xvfi}. Such approximations often assume motion linearity, and hence are prone to errors in non-linear motion scenarios. Various efforts have been made to alleviate this issue, including the use of depth information~\cite{bao2019depth}, higher order motion models~\cite{xu2019quadratic, liu2020enhanced}, and adaptive forward warping~\cite{niklaus2020softmax}. A second group of methods~\cite{liu2017video, xue2019video, park2020bmbc, zhang2020flexible, huang2020rife, park2021asymmetric} have been developed to improve approximation by directly predicting intermediate flows. These approaches typically employ a coarse-to-fine architecture, which supports a larger receptive field for capturing large motions. In all of the above methods, the predicted flows correspond to a one-to-one pixel mapping, which inherently limits the ability to capture complex motions.

\noindent\textbf{Kernel-based VFI.} In these methods, various convolution kernels~\cite{adaconv, niklaus2017video, lee2020adacof, shi2020video, ding2021cdfi, cheng2021multiple, chen2021pdwn,gui2020featureflow, long2016learning, choi2020channel, kalluri2020flavr} are learned as a basis for synthesizing interpolated pixels. Earlier approaches~\cite{adaconv, niklaus2017video} predict a fixed-size kernel for each output location, which is then convolved with co-located input pixels. This limits the magnitude of captured motions to the kernel size used, while more memory and computational capacity are required when larger kernel sizes are adopted. To overcome this problem, deformable convolution (DefConv)~\cite{dai2017deformable} was adapted to VFI in AdaCoF~\cite{lee2020adacof}, which allows kernels to be convolved with any input pixels pointed by local offset vectors. This can be considered as \textit{multi-interflows}, representing a many-to-one mapping. Further improvements to AdaCoF have been achieved by allowing space-time sampling~\cite{shi2020video}, feature pyramid warping~\cite{ding2021cdfi}, and using a coarse-to-fine architecture~\cite{chen2021pdwn}.

\subsection{Dynamic Texture Synthesis}

Dynamic textures (e.g. water, fire, leaves blowing in the wind etc.) generally exhibit high spatial frequency energy alongside temporal stochasticity, with inter-frame motions irregular in both the spatial and temporal domains. Classic synthesis methods rely on mathematical models such as Markov random fields~\cite{wei2000fast} and auto-regressive moving average model~\cite{doretto2003dynamic} to capture underlying motion characteristics. More recently, deep learning techniques, in particular 3D CNNs and GAN-based training~\cite{gatys2015texture, yang2016stationary, xie2019learning,wang2021conditional,yang2021spatiotemporal}, have been adopted to achieve more realistic synthesis results. It should be noted that both dynamic texture synthesis and VFI require accurate modeling of spatio-temporal characteristics. However the techniques developed specifically for texture synthesis have not yet been fully exploited in VFI methods. This is a focus of our work.

\begin{figure*}[ht]
\subfloat[MIFNet] {\includegraphics[width=0.68\linewidth]{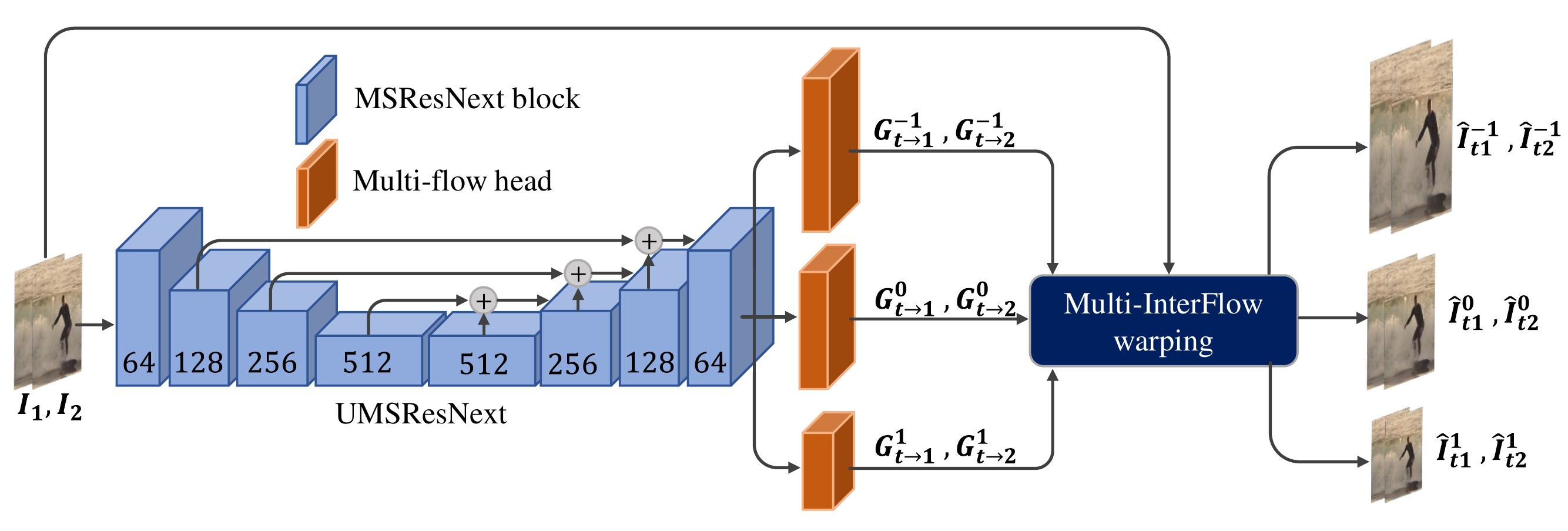}}\;\!\!
\hspace{8mm}
\subfloat[Multi-flow head] {\includegraphics[width=0.23\linewidth]{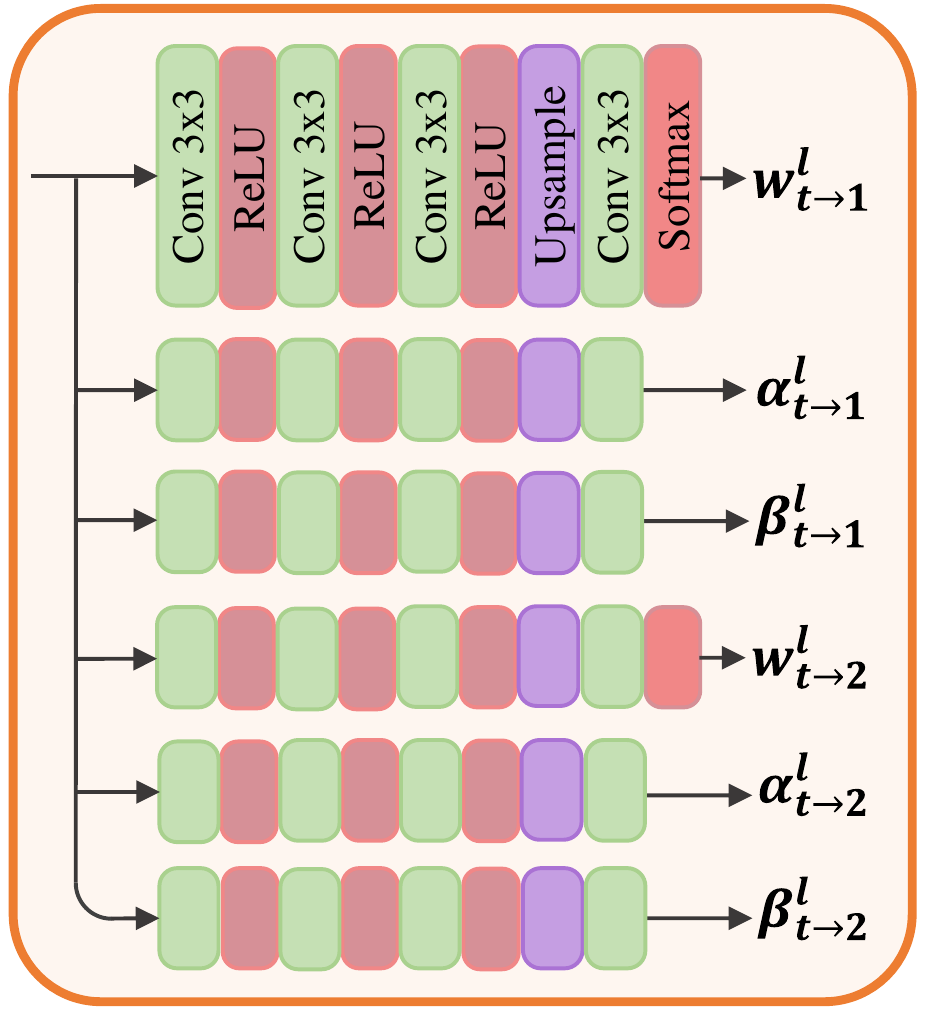}}\;\!\!
\vspace{-3mm}
\caption{\label{fig:mifnet}
Illustration of the MIFNet. (a) The overall architecture of MIFNet, with a U-Net style backbone and multi-flow estimation heads at three scales. (b) The convolutional layers inside the multi-flow head at each scale.}
\vspace{-4mm}
\end{figure*}
\begin{figure}[ht]
\begin{center}
\includegraphics[width=0.9\linewidth]{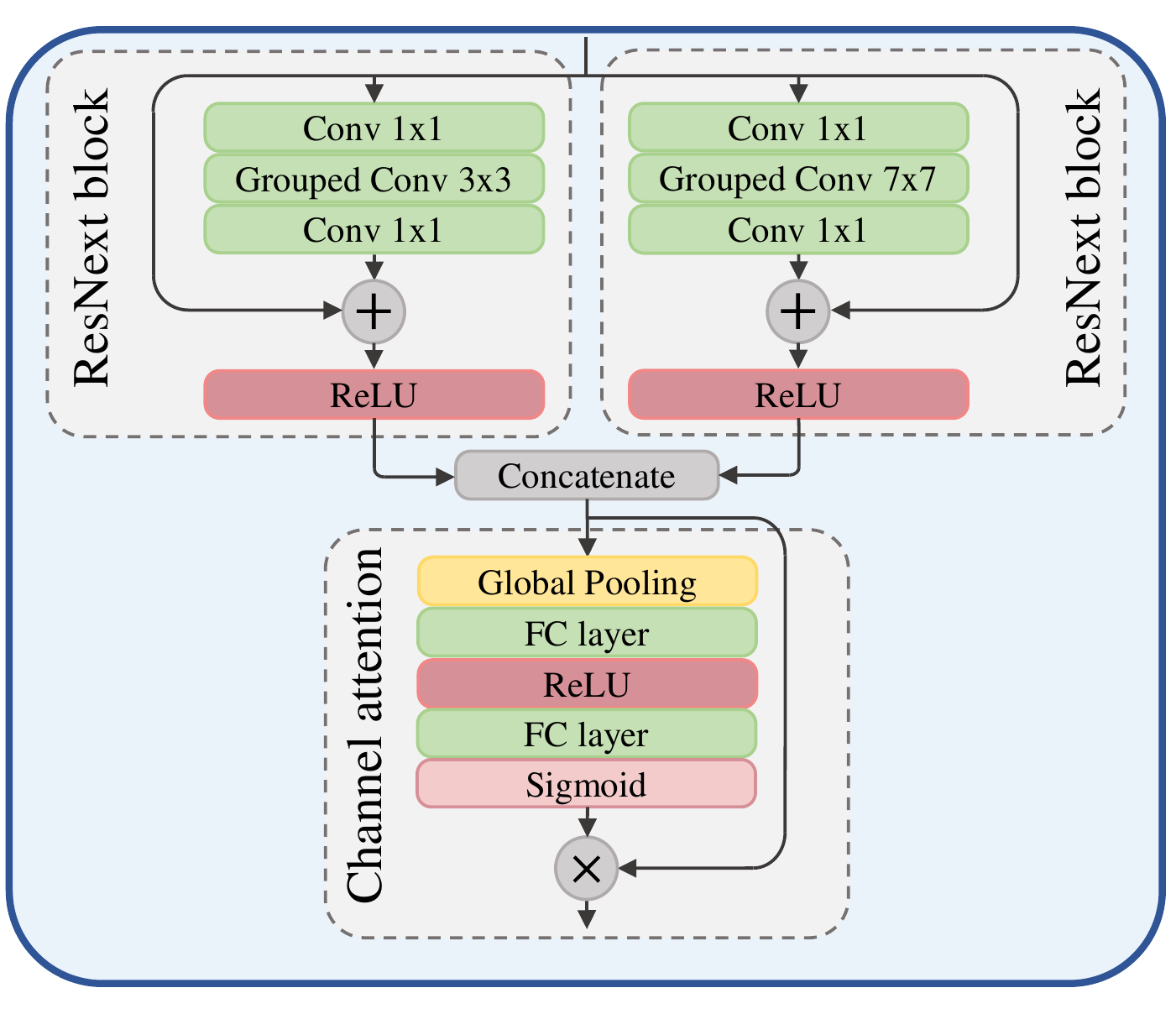}
\end{center}
\vspace{-6mm}
\caption{\label{fig:msresnextblock}
Illustration of the MSResNext block, which consists of two ResNext branches with different kernel sizes, followed by a channel attention module.}
\vspace{-4mm}
\end{figure}

\section{Proposed Method: ST-MFNet}

Figure~\ref{fig:overview} shows the architecture of ST-MFNet. While conventionally VFI is formulated as generating the intermediate frame $I_t$ ($t=1.5$) between two given consecutive frames $I_1, I_2$, we instead employ two more frames $I_0, I_3$ to improve the modeling of motion dynamics. Given the frames $I_0, I_1, I_2, I_3$, our model first processes $I_1, I_2$ in two branches. The Multi-InterFlow Network (MIFNet) branch estimates the multi-scale multi-flows from $I_t$ to $I_1, I_2$, where the many-to-one pixel correspondence allows complex transformation, benefiting interpolation of highly complex motion, such as dynamic textures (e.g. water, fire etc.). As the fixed receptive field of MIFNet may lead to limited ability to capture large motion, we also included the Bi-directional Linear Flow Network (BLFNet) branch to approximate one-to-one optical flows from $I_1, I_2$ to $I_t$ using a coarse-to-fine approach, enhancing large motion capturing. The input frames are warped based on the flows generated by MIFNet and BLFNet, and then fused by the Multi-Scale Fusion module to obtain an intermediate result $\tilde{I}_t$. This multi-branch structure combines the advantages of both single-flow and multi-flow based methods and was found to offer enhanced interpolation performance. In the second stage, $\tilde{I}_t$ is combined with all the inputs $I_0, I_1, I_2, I_3$ in temporal order and fed into the Texture Enhancement Network (TENet), which captures longer-range dynamic and generates residual signals for the final output.

\subsection{Multi-InterFlow Network} \label{sec:method_1}

\noindent\textbf{Multi-InterFlow warping.} For self-completeness, we first briefly describe the multi-interflow warping operation~\cite{lee2020adacof}. Given two images $I_A, I_B$ with size $H\times W$, conventional optical flow $F_{A\rightarrow B} = (\mathbf{f}_x, \mathbf{f}_y)$ from $I_A$ to $I_B$ specifies the x- and y-components of pixel-wise offset vectors, where $\mathbf{f}_x, \mathbf{f}_y\in \mathbb{R}^{H\times W}$. The pixel value at each location $(x,y)$ of the corresponding backwarped~\cite{jaderberg2015spatial} $\hat{I}_A$ is defined as
\begin{gather}
    \hat{I}_A(x,y) = I_B(x+\mathbf{f}_x(x,y), y+\mathbf{f}_y(x,y))
\end{gather}
where the values at non-integer grid locations are obtained via bilinear interpolation. The multi-interflow proposed in~\cite{lee2020adacof} can be defined as $G_{A\rightarrow B} = (\boldsymbol{\alpha}, \boldsymbol{\beta}, \mathbf{w})$, but now $\boldsymbol{\alpha}, \boldsymbol{\beta} \in \mathbb{R}^{H\times W\times N}$ represent a collection of the x- and y-components of $N$ flow vectors respectively and $\mathbf{w}\in [0,1]^{H\times W\times N}$ is their weighting kernels ($\sum_{i=1}^N \mathbf{w}(x,y,i)=1$). That is, for each location $(x,y)$, $G_{A\rightarrow B}$ contains $N$ flow vectors and $N$ weights. The corresponding warping is defined as follows.
\vspace{-10pt}

\small
\begin{equation}
\label{eqn2}
    \hat{I}_A(x,y) = \sum_{i=1}^N \mathbf{w}(x,y,i) \cdot I_B(x+\boldsymbol{\alpha}(x,y,i), y+\boldsymbol{\beta}(x,y,i))
\end{equation}
\normalsize
\vspace{-10pt}

\noindent Such multi-flow warping corresponds to a many-to-one mapping, which allows flexible sampling of source pixels, enabling the capture of more complex motions. 

Given input frames $I_1, I_2$, the MIFNet predicts the multi-interflows $\{G^l_{t\rightarrow 1}, G^l_{t\rightarrow 2}\}$ from the intermediate frame $I_t$ to the inputs at three scale levels: $l=-1,0,1$, where $l=i$ means spatial down-sampling by $2^i$ (i.e. $l=-1$ denotes up-sampling), so that re-sampled inputs $I^l_{1},I^l_{2}$ can be warped to time $t$ using Equation (\ref{eqn2}) to produce $\hat{I}^l_{t1},\hat{I}^l_{t2}$ respectively. 

\noindent\textbf{Architecture.} Figure~\ref{fig:mifnet}~(a) shows the architecture of the MIFNet. In order to capture pixel movements at multiple scales, we devise a U-Net style feature extractor, U-MultiScaleResNext (UMSResNext), consisting of eight MSResNext blocks (shown in Figure~\ref{fig:msresnextblock}). Each MSResNext block employs two ResNext blocks~\cite{xie2017aggregated} in parallel with different kernel sizes in the middle layer, 3$\times$3 and 7$\times$7, which further increases the network \textit{cardinality}~\cite{xie2017aggregated,ma2020cvegan}. The outputs of these two ResNext blocks are then concatenated and connected to a channel attention module~\cite{Hu_2018_CVPR}, which learns adaptive weighting of the feature maps extracted by the two ResNext blocks. Such feature selection mechanism has also been found to enhance motion modeling~\cite{choi2020channel, kalluri2020flavr}. In UMSResNext, the up-sampling operation is performed by replacing the k$\times$k grouped convolutions in the middle layer with (k+1)$\times$(k+1) grouped transposed convolutions.

The features extracted by UMSResNext are then passed to the multi-flow heads for multi-interflow prediction. In contrast to \cite{lee2020adacof}, multi-flows here are predicted at various scales $l=-1,0,1$, and occlusion maps are not generated (occlusion is handled by the BLFNet). As shown in Figure~\ref{fig:mifnet} (b), each multi-flow head contains 6 sub-branches, predicting the x-, y-components ($\boldsymbol{\alpha},\boldsymbol{\beta}$) and the kernel weights ($\mathbf{w}$) of ${G^l_{t\rightarrow 1},G^l_{t\rightarrow 2}}$. The predicted flows are then used to backwarp the inputs $I_1,I_2$ at corresponding scales using Equation~(\ref{eqn2}). Here a bilinear filter is used for down-sampling input frames, and an 8-tap filter originally designed for sub-pixel motion estimation~\cite{sullivan2012overview} is employed for up-sampling. The down-sampled scale used here encourages the motion search in a larger region, while the up-sampled scale allows the motion vectors to point to finer sub-pixel locations, increasing the precision of multi-flow warping.

\subsection{Bi-directional Linear Flow Network}

To improve large motion interpolation, bi-directional flows $F_{1\rightarrow 2}, F_{2\rightarrow 1}$ between inputs $I_1, I_2$ are also predicted using a pre-trained flow estimator~\cite{sun2018pwc}, which is based on a coarse-to-fine architecture. The intermediate flows are then linearly approximated as follows.
\begin{equation}
    F_{1\rightarrow t} = 0.5F_{1\rightarrow 2} \quad F_{2\rightarrow t} = 0.5F_{2\rightarrow 1}
\end{equation}
According to the intermediate flows, the frames $I_1, I_2$ are forward warped using the efficient softsplat operator~\cite{niklaus2020softmax}, which learns occlusion-related softmax-alike weighting of reference pixels in the forward warping process. Another advantage of softsplat is that it is differentiable, allowing the flow estimator to be end-to-end optimized. Finally, BLFNet branch outputs warped frames $\hat{I}_{t1}^\text{soft}, \hat{I}_{t2}^\text{soft}$. The employment of the BLFNet branch was found to be essential for handling large motion and occlusion and improving the overall capacity of the proposed model.

\subsection{Multi-Scale Fusion Module}
The Multi-Scale Fusion Module is employed to produce an intermediate interpolation result using the frames warped at multiple scales in the previous steps. Here we adopt the GridNet~\cite{fourure2017residual} architecture due to its superior performance on fusing multi-scale information~\cite{niklaus2018context, niklaus2020softmax}. The GridNet is configured here to have 4 columns and 3 rows, with the first, second and third rows corresponding to scales of $l=-1,0,1$ respectively. The first and third rows take $\{\hat{I}^{-1}_{t1}, \hat{I}^{-1}_{t2}\}$ and $\{\hat{I}^{1}_{t1}, \hat{I}^{1}_{t2}\}$ as inputs, while the second row takes $\{\hat{I}^{0}_{t1}, \hat{I}^{0}_{t2}, \hat{I}_{t1}^\text{soft}, \hat{I}_{t2}^\text{soft}\}$, where $\{\cdot\}$ denotes channel-wise concatenation. Finally, this module outputs the intermediate result $\tilde{I}_t$ at the original spatial resolution ($l=0$).

\begin{figure}[t]
\begin{center}
   \includegraphics[width=0.9\linewidth]{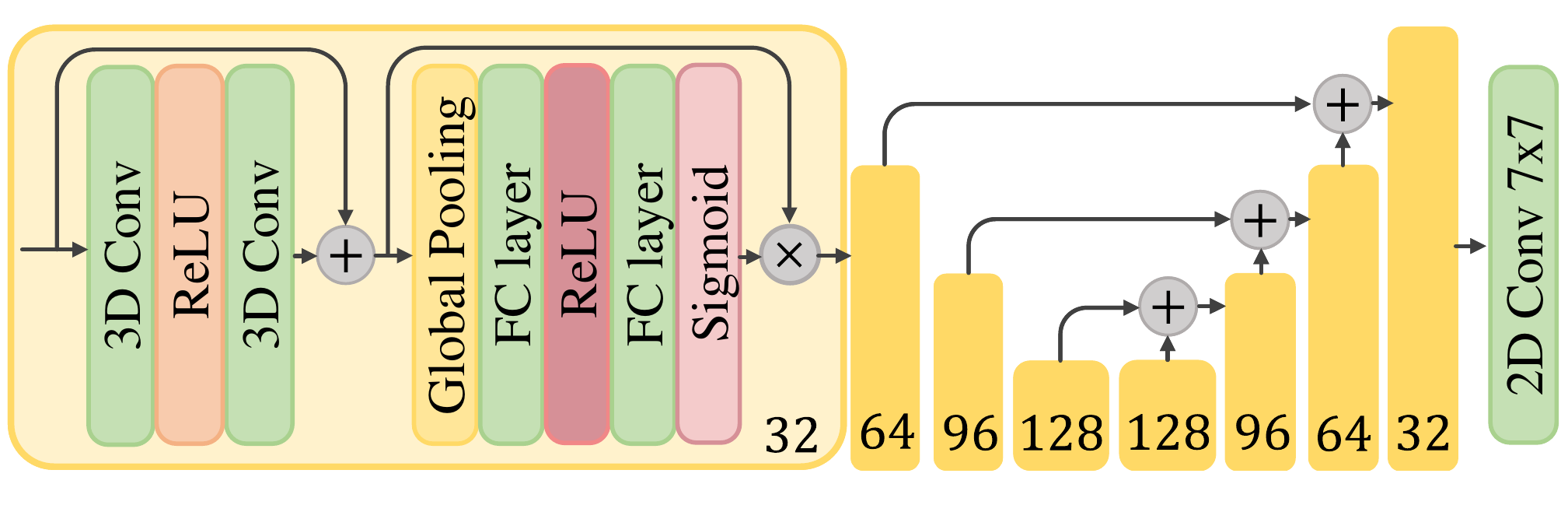}
\end{center}
\vspace{-7mm}
\caption{\label{fig:tenet}
The architecture of the Texture Enhancement Network.}
\vspace{-4.5mm}
\end{figure}

\subsection{Texture Enhancement Network}

At the end of the first stage, the output of the Multi-Scale Fusion module, $\tilde{I}_t$, is concatenated with four original inputs to form $\{I_0,I_1,\tilde{I}_t,I_2,I_3\}$, which are then fed into the Texture Enhancement Network (TENet). Including additional frames here allows better modeling of higher-order motions and also provides more information on longer-term spatio-temporal characteristics. Motivated by recent work in dynamic texture synthesis~\cite{xie2019learning, yang2021spatiotemporal}, where spatio-temporal filtering was found to be effective for generating coherent video textures, we integrate a 3D CNN for texture enhancement. This CNN architecture (shown in Figure~\ref{fig:tenet}) is a modified version of the network developed in \cite{kalluri2020flavr}, but with reduced layer widths. This is based on the consideration that the intermediately warped frame $\tilde{I}_t$ has already been produced which is relatively close to the target. It is different from the original scenario in \cite{kalluri2020flavr}, where the network is expected to directly synthesize the interpolated output using the four original input frames. Finally, the TENet is expected to output a residual signal containing textural difference between $\tilde{I}_t$ and the target frame, which contributes to the final output of ST-MFNet.


\subsection{Loss Functions}\label{sec:loss}

We trained two versions of ST-MFNet in this work. For the distortion oriented model, a Laplacian pyramid loss~\cite{bojanowski2017optimizing} ($\mathcal{L}_{lap}$) was used as the objective function. This model was further fine-tuned using an ST-GAN based perceptual loss ($\mathcal{L}_{p}$) to obtain the perceptually optimized version.

\noindent\textbf{Laplacian pyramid loss.} ST-MFNet was trained end-to-end by matching its output $I^{out}_t$ with the ground-truth intermediate frame $I^{gt}_t$ using the Laplacian pyramid loss~\cite{bojanowski2017optimizing}, which has been previously used for VFI in~\cite{niklaus2018context,niklaus2020softmax,liu2020enhanced}. The loss function is defined below.
\begin{equation}
    \mathcal{L}_{lap} = \sum_{s=1}^{S} 2^{s-1} \norm{L^s(I^{out}_t)-L^s(I^{gt}_t)}_1 \label{l_lap}
\end{equation}
Here $L^s(I)$ denotes the $s$\textsuperscript{th} level of the Laplacian pyramid of an image $I$, and $S$ is the maximum level. 

\noindent\textbf{Spatio-temporal adversarial loss.}
To further improve the perceptual quality of the ST-MFNet output, we also trained our model using the Spatio-Temporal Generative Adversarial Networks (ST-GAN) training methodology~\cite{yang2021spatiotemporal}. Different from the conventional GAN~\cite{goodfellow2020generative} focusing on a single image, the discriminator $D$ here also processes adjacent video frames which improves temporal consistency. This is key for video frame interpolation. The architecture of the discriminator is provided in Appendix~\ref{sec:dis}. This discriminator was trained with the following loss.
\begin{equation}
    \mathcal{L}_D = -\log (1-D(I^{out}_t, I_1, I_2)) - \log (D(I_t^{gt}, I_1, I_2))
\end{equation}
The corresponding adversarial loss for the generator (ST-MFNet) is given below.
\begin{equation}
    \mathcal{L}_{adv} = -\log (D(I^{out}_t, I_1, I_2))
\end{equation}
This is then combined with the Laplacian pyramid loss to form the perceptual loss for ST-MFNet fine-tuning,
\begin{align} 
    \mathcal{L}_{p} = \mathcal{L}_{lap} + \lambda\mathcal{L}_{adv} \label{l_p}
\end{align}
where $\lambda$ is a weighting hyper-parameter that controls the perception-distortion trade-off~\cite{blau2018perception}.

\section{Experimental Setup}\label{sec:expsetup}
\noindent\textbf{Implementation details.}
In our implementation, we set the number of flows $N=25$ (the default value in \cite{lee2020adacof}) for the MIFNet branch. The maximum level $S$ for $\mathcal{L}_{lap}$ was set to 5, and the weighting hyper-parameter $\lambda=100$. We used the AdaMax optimizer~\cite{kingma2014adam} with $\beta_1=0.9,\beta_2=0.999$. The learning rate was set to 0.001 and reduced by a factor of 0.5 whenever the validation performance stops improving for 5 epochs. The pre-trained flow estimator~\cite{sun2018pwc} in the BLFNet branch was frozen for the first 60 epochs and then fine-tuned for 10 more epochs to further improve VFI performance. The network was trained for a total number of 70 epochs using a batch size of 4. All training and evaluation were executed with NVIDIA P100~\cite{BC4} and RTX 3090 GPUs.

\noindent\textbf{Training datasets.} We used the training split of Vimeo-90k (septuplet) dataset~\cite{xue2019video} which contains 91,701 frame septuplets (448$\times$256). As Vimeo-90k was produced with constrained motion magnitude and complexity, to further enhance the VFI performance on large motion and dynamic textures, we used an additional dataset, BVI-DVC~\cite{ma2020bvi}, which covers a wide range of texture/motion types, frame rates (24 to 120 FPS) and spatial resolutions (2160p, 1080p, 540p, and 270p). We randomly sampled 12800, 6400, 800, 800 septuplets from videos at each of these resolutions respectively, leaving out a subset of frames for validation. We augmented all septuplets from both datasets by randomly cropping 256$\times$256 patches and performing flipping and temporal order reversing. This resulted in more than 100,000 septuplets of 256$\times$256 patches. In each septuplet, the 1\textsuperscript{st}, 3\textsuperscript{rd}, 5\textsuperscript{th} and 7\textsuperscript{th} frames were used as inputs and the 4\textsuperscript{th} as the training target. The test split of Vimeo-90k together with unused subset of BVI-DVC was utilized as the validation set for hyper-parameter tuning and training monitoring.

\noindent\textbf{Evaluation dataset.}
Since our model takes four frames as input, the evaluation dataset should be able to provide frame quintuplets $I_0,I_1,I_t^{gt},I_2,I_3$ ($t=1.5$). In this work, we used the test quintuplets in \cite{xu2019quadratic}, which were extracted from the UCF-101~\cite{soomro2012ucf101} (100 quintuplets) and DAVIS~\cite{perazzi2016benchmark} (2847 quintuplets) datasets. We also evaluated on the SNU-FILM dataset~\cite{choi2020channel}, which specifies a list of 310 triplets at four motion magnitude levels. As original sequences are provided in the SNU-FILM dataset, we extended its pre-defined test triplets into quintuplets for the evaluation here. 

\begin{figure*}[t]
	\begin{center}
		\subfloat[Overlay]
		{\includegraphics[width=0.117\linewidth]{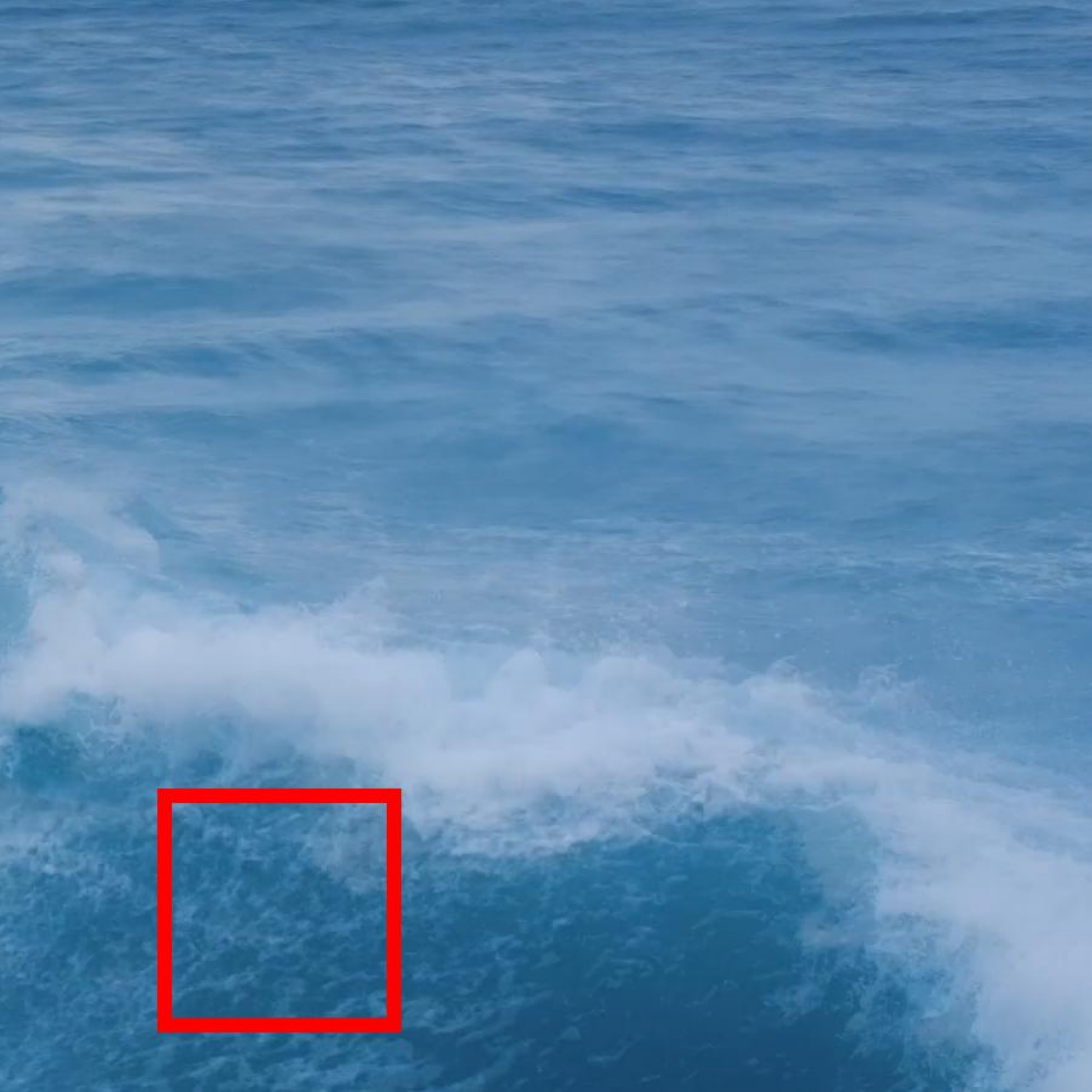}}\,
		\subfloat[GT]
		{\includegraphics[width=0.117\linewidth]{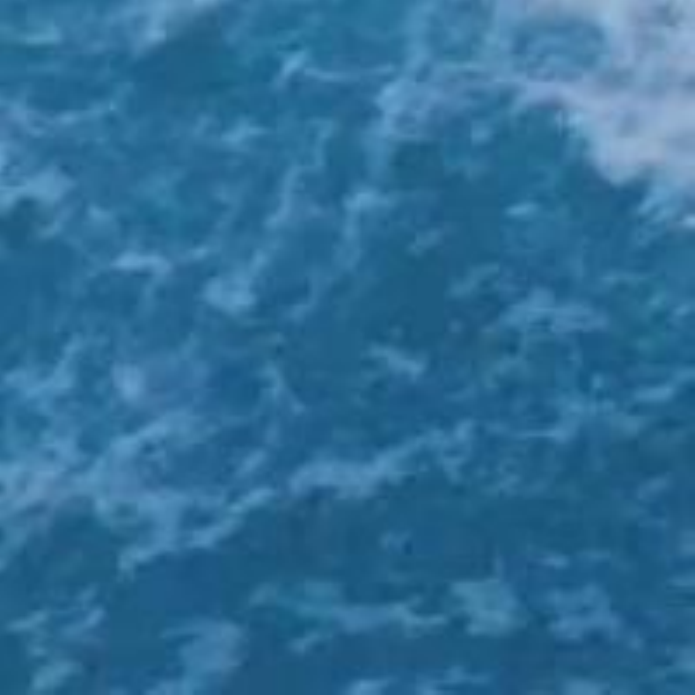}}\,
		\subfloat[w/o MIFNet]
		{\includegraphics[width=0.117\linewidth]{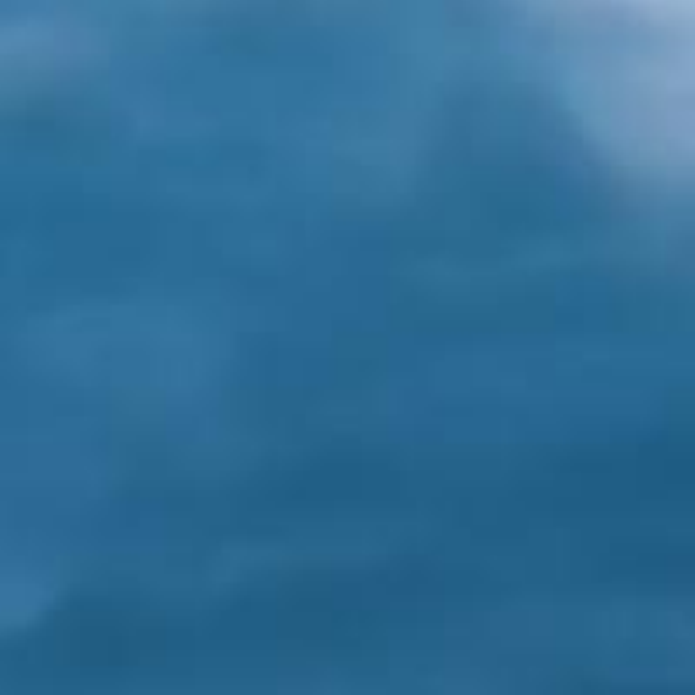}}\,
		\subfloat[w/ MIFNet]
		{\includegraphics[width=0.117\linewidth]{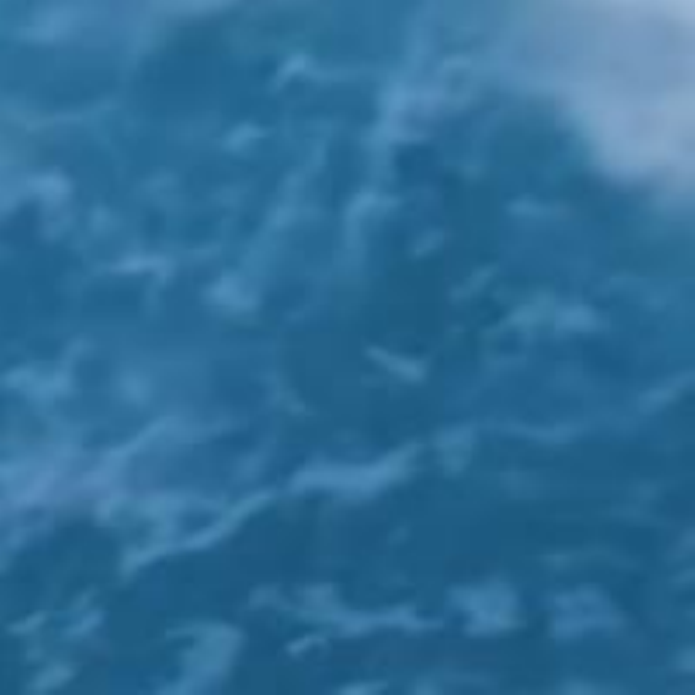}}\,\,
		\subfloat[Overlay]
		{\includegraphics[width=0.117\linewidth]{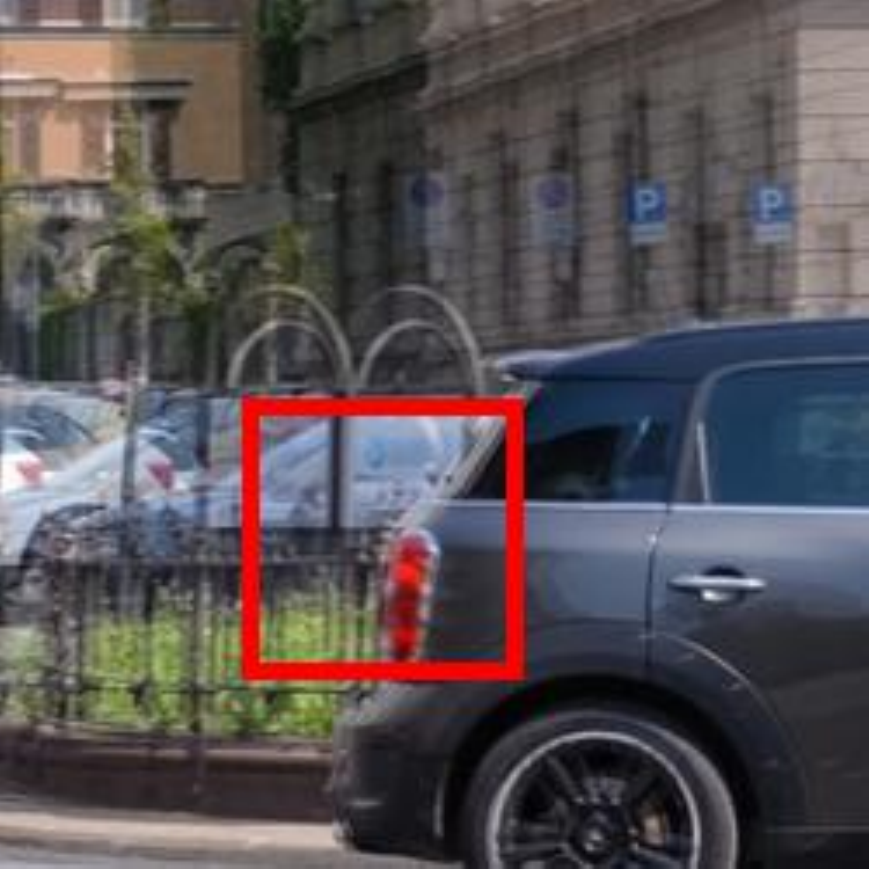}}\,
		\subfloat[GT]
		{\includegraphics[width=0.117\linewidth]{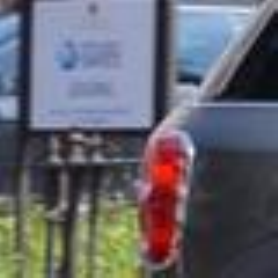}}\,
		\subfloat[w/o BLFNet]
		{\includegraphics[width=0.117\linewidth]{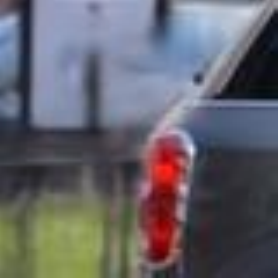}}\,
		\subfloat[w/ BLFNet]
		{\includegraphics[width=0.117\linewidth]{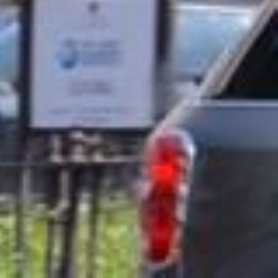}}\\
		\subfloat[Overlay]
		{\includegraphics[width=0.117\linewidth]{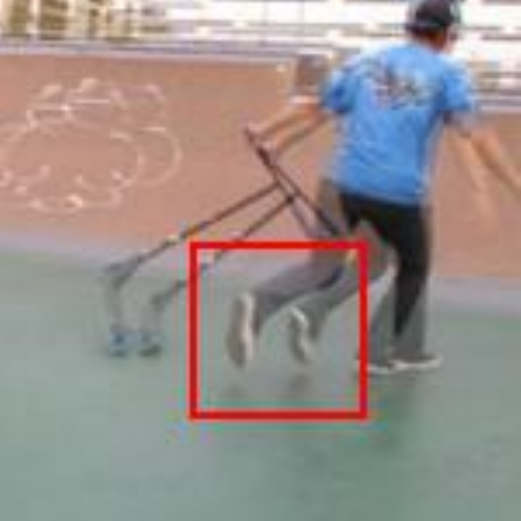}}\,
		\subfloat[GT]
		{\includegraphics[width=0.117\linewidth]{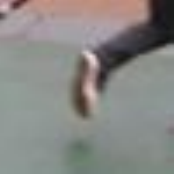}}\,
		\subfloat[U-Net]
		{\includegraphics[width=0.117\linewidth]{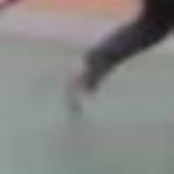}}\,
		\subfloat[UMSResNext]
		{\includegraphics[width=0.117\linewidth]{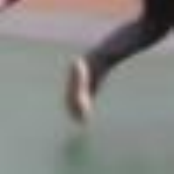}} \,\,
		\subfloat[Overlay]
		{\includegraphics[width=0.117\linewidth]{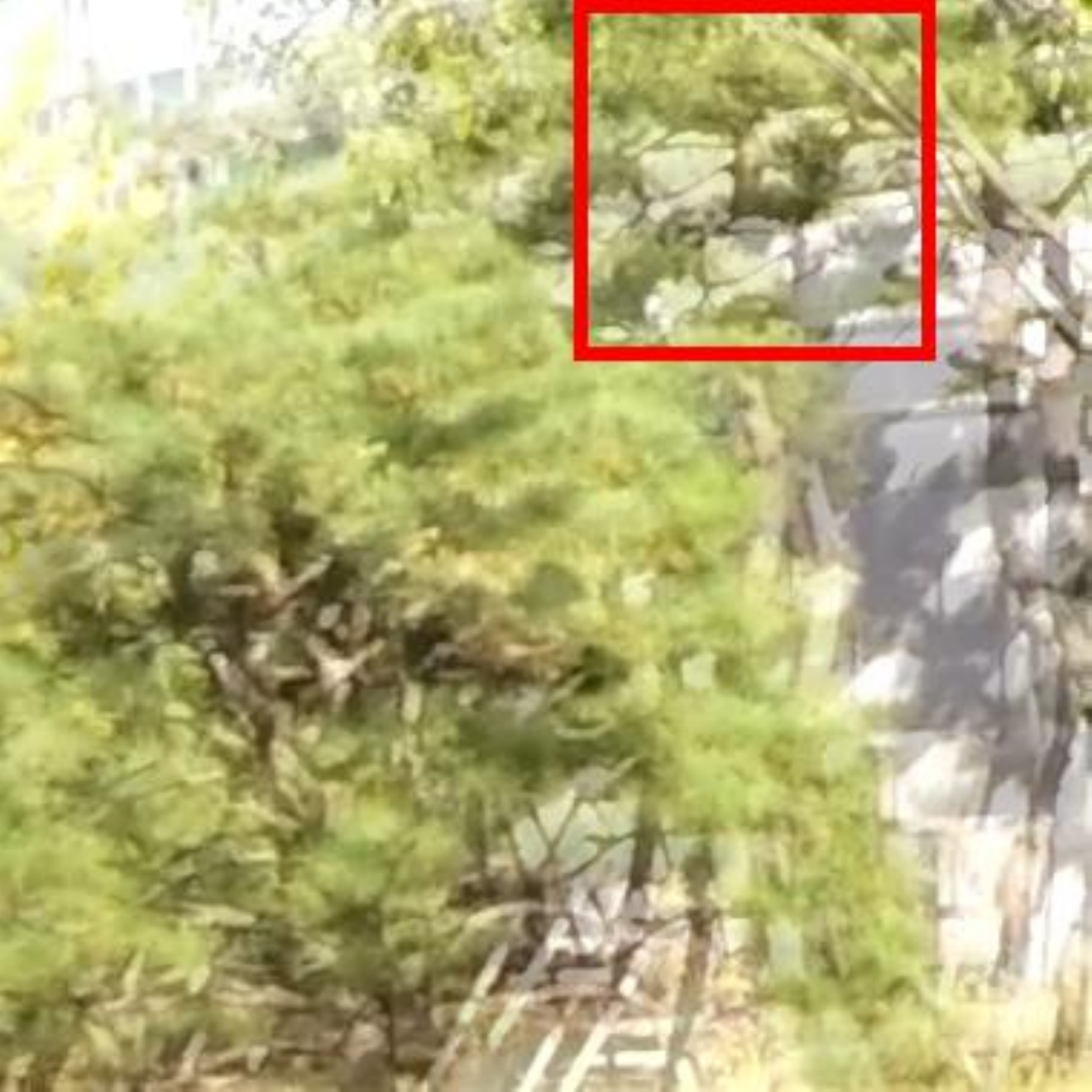}}\,
		\subfloat[GT]
		{\includegraphics[width=0.117\linewidth]{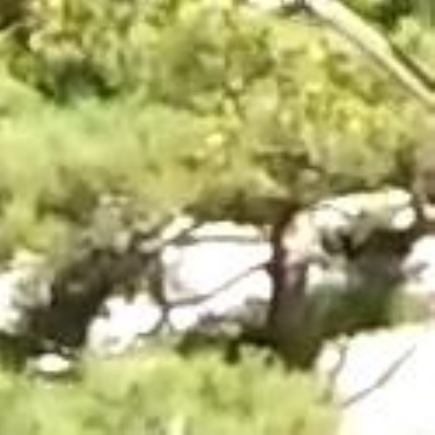}}\,
		\subfloat[w/o TENet]
		{\includegraphics[width=0.117\linewidth]{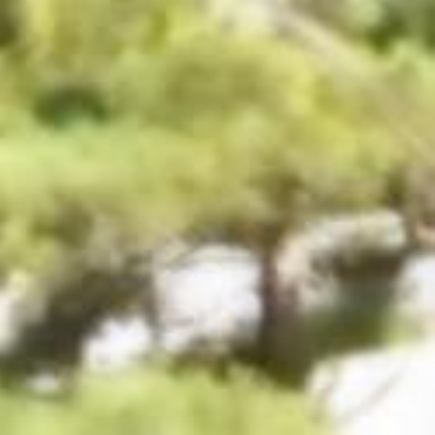}}\,
		\subfloat[w/ TENet]
		{\includegraphics[width=0.117\linewidth]{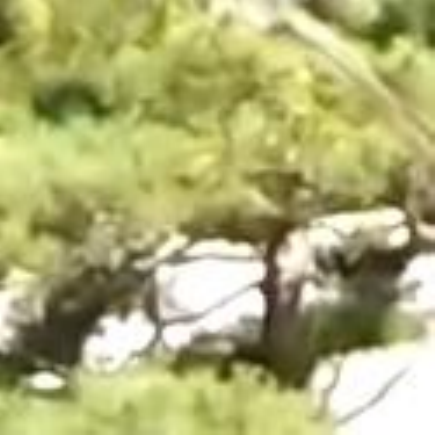}}\\
        \subfloat[Overlay]
    	{\includegraphics[width=0.16\linewidth]{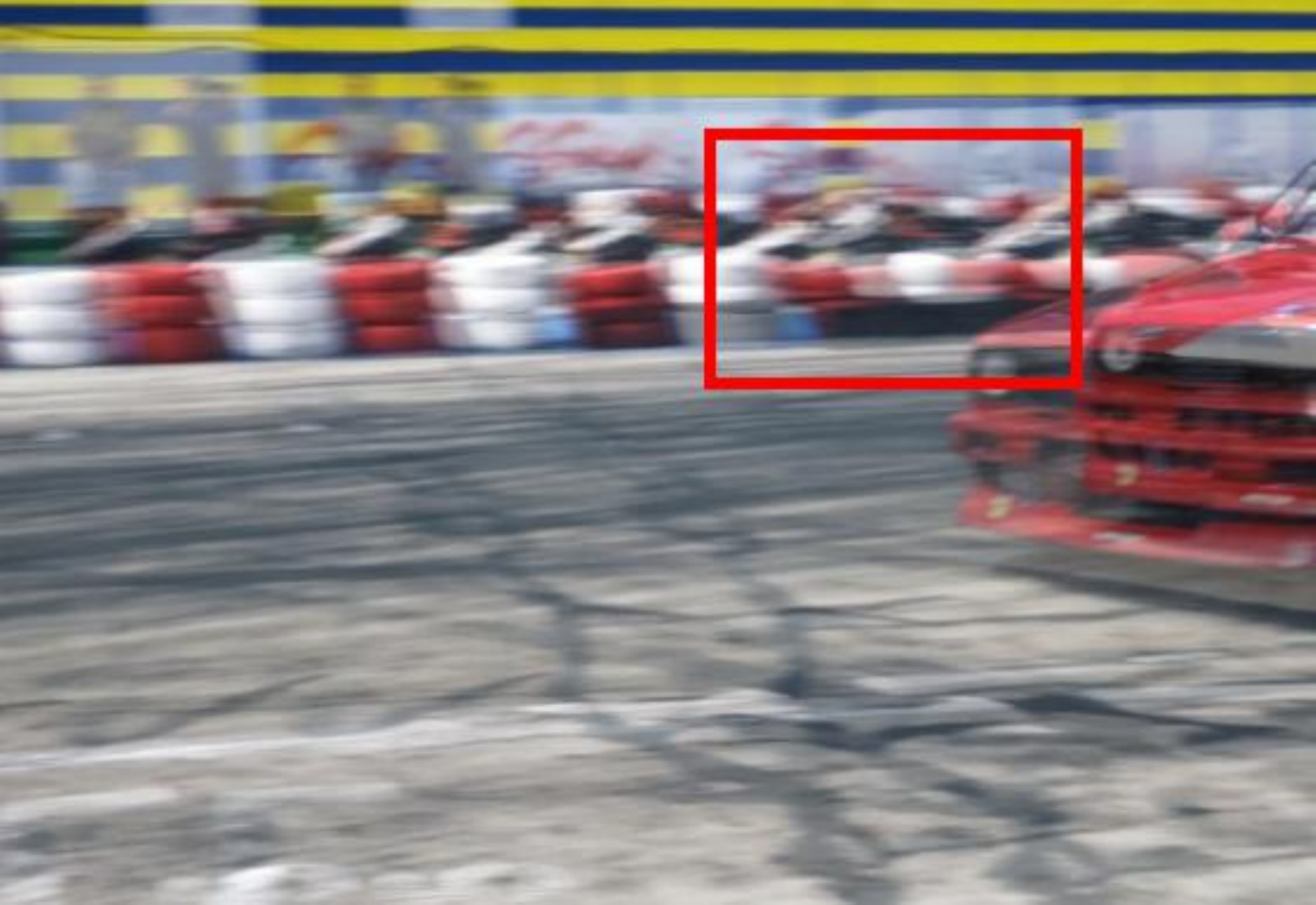}} \,
    	\subfloat[GT]
    	{\includegraphics[width=0.16\linewidth]{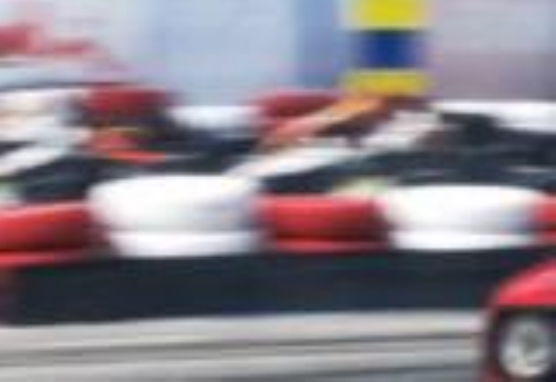}} \,
    	\subfloat[Ours-$\mathcal{L}_{lap}$]
    	{\includegraphics[width=0.16\linewidth]{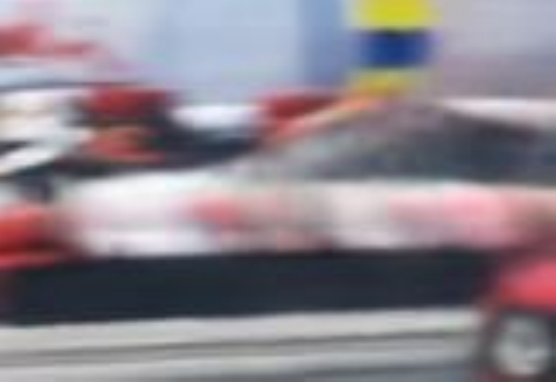}} \,
    	\subfloat[TGAN]
    	{\includegraphics[width=0.16\linewidth]{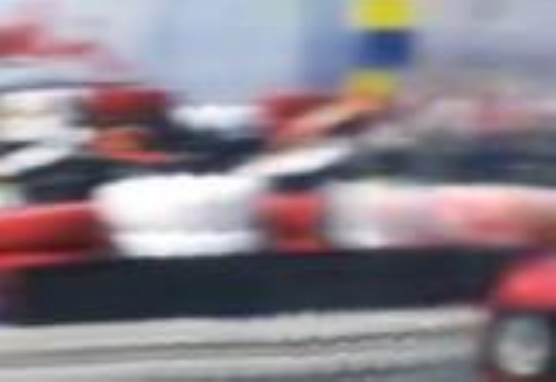}} \,
    	\subfloat[FIGAN]
    	{\includegraphics[width=0.16\linewidth]{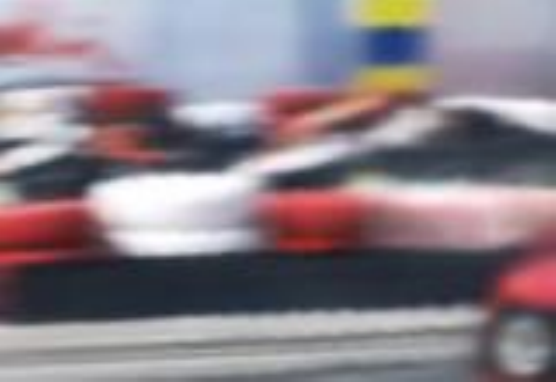}} \,
    	\subfloat[Ours-$\mathcal{L}_{p}$]
    	{\includegraphics[width=0.16\linewidth]{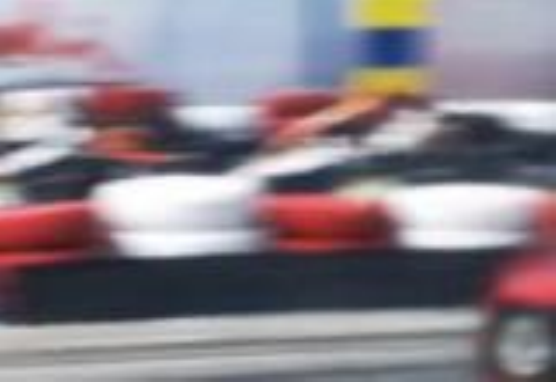}}
 \end{center}
\vspace{-5mm}
\caption{Qualitative results interpolated by different variants of our method. Here ``Overlay" means the overlaid adjacent frames. Figures (a)-(d): w/ MIFNet vs w/o MIFNet; figures (e)-(h): w/ BLFNet vs w/o BLFNet; figures (i)-(j): UMSResNext vs U-Net; figures (m)-(p): w/ TENet vs w/o TENet; figures (q)-(v): comparison of different GANs.}
\label{fig:ablation}
\vspace{-3mm}
\end{figure*}

To further test interpolation performance on various texture types, we developed a new test set, VFITex, which contains twenty 100-frame UHD or HD videos at 24, 30 or 50 FPS, collected from the Xiph~\cite{montgomery3xiph}, Mitch Martinez Free 4K Stock Footage~\cite{mitch}, UVG database~\cite{mercat2020uvg} and the Pexels website~\cite{pexels}. This dataset covers diverse textured scenes, including crowds, flags, foliage, animals, water, leaves, fire and smoke. Based on the computational capacity available, we center-cropped HD patches from the UHD sequences, preserving the original UHD characteristics. All frames in each sequence were used for evaluation, totaling 940 quintuplets. More details of the training and evaluation datasets and their license information are provided in Appendix~\ref{sec:license}.

\noindent\textbf{Evaluation Methods.}
Two most commonly used quality metrics, PSNR and SSIM~\cite{wang2004image}, were employed here for objective assessment of the interpolated content. We note that these metrics do not always correlate well with video quality as perceived by a human observer~\cite{hore2010image, kalluri2020flavr}. Therefore, a more perceptually motivated metric, LPIPS~\cite{zhang2018unreasonable}, were used. Furthermore, in order to directly compare the perceptual quality of the video frames interpolated by our method and the benchmark references, a user study was conducted based on a psychophysical experiment. The details of the user study are described in Section~\ref{sec:qualitative}.
\begin{table}[t]
\resizebox{\columnwidth}{!}{\begin{tabular}{llll}
\toprule
          & \multicolumn{1}{c}{UCF101} & \multicolumn{1}{c}{DAVIS} & \multicolumn{1}{c}{VFITex} \\ \hline
Ours-\textit{w/o BLFNet} & 33.218/0.970 & 27.767/0.881 & 28.498/0.915 \\
Ours-\textit{w/o MIFNet} & 33.202/0.969 & 27.886/0.889 & 28.357/0.911 \\ 
Ours-\textit{w/o TENet} & 32.895/0.970 & 27.484/0.880 & 28.241/0.910 \\ 
Ours-\textit{unet} & 33.378/0.970 & 28.096/0.892 & 28.898/0.925 \\
\hline
Ours & 33.384/0.970 & 28.287/0.895 & 29.175/0.929 \\
\bottomrule
\end{tabular}}
\vspace{-2.5mm}
\caption{Ablation study results (PSNR/SSIM) for ST-MFNet.}
\label{tab:ablation}
\vspace{-5mm}
\end{table}

\begin{table*}[t]
	\begin{center}
		\resizebox{\linewidth}{!}{
			\begin{tabular}{lccccccccc}
				\toprule
				& \multirow{2}[1]{*}{UCF101} & \multirow{2}[1]{*}{DAVIS} & \multicolumn{4}{c}{SNU-FILM} & \multirow{2}[1]{*}{VFITex} & \multirow{2}[2]{*}{\makecell{RT \\ (sec)}} & \multirow{2}[2]{*}{\makecell{\#P \\ (M)}} \\
				\cmidrule(l{5pt}r{5pt}){4-7}
				& & &Easy&Medium&Hard&Extreme& & \\
				\midrule
				DVF~\cite{liu2017video} & 32.251/0.965  & 20.403/0.673 & 27.528/0.876 & 24.091/0.817 & 21.556/0.760 & 19.709/0.705 & 19.946/0.709 & 0.157 & 3.82\\
				SuperSloMo~\cite{jiang2018super} & 32.547/0.968 & 26.523/0.866  & 36.255/0.984 & 33.802/0.973 & 29.519/0.930 & 24.770/0.855 & 27.914/0.911 & 0.107 & 39.61\\
				SepConv~\cite{niklaus2017video} & 32.524/0.968 & 26.441/0.853 & 39.894/0.990 & 35.264/0.976 & 29.620/0.926 & 24.653/0.851 & 27.635/0.907 & 0.062 & 21.68\\
				DAIN~\cite{bao2019depth} & \underline{32.524}/\underline{0.968} & 27.086/0.873 & 39.280/0.989 & 34.993/0.976 & 29.752/0.929 & 24.819/0.850 & 27.314/0.909 & 0.896 & 24.03\\
				BMBC~\cite{park2020bmbc} & \underline{32.729}/\underline{0.969} & 26.835/0.869 & 39.809/0.990 & 35.437/0.978 & 29.942/0.933 & 24.715/0.856 & 27.337/0.904 & 1.425 & 11.01\\
				AdaCoF~\cite{lee2020adacof} & \underline{32.610}/\underline{0.968} & 26.445/0.854 & 39.912/0.990 & 35.269/0.977 & 29.723/0.928 & 24.656/0.851 & 27.639/0.904 & 0.051 & 21.84\\
				FeFlow~\cite{gui2020featureflow} & 32.520/0.967 & 26.555/0.856 & 39.591/0.990 & 35.014/0.977 & 29.466/0.928 & 24.607/0.852 & OOM & 1.385 & 133.63\\
				CDFI~\cite{ding2021cdfi} & \underline{32.653}/\underline{0.968} & 26.471/0.857 & 39.881/0.990 & 35.224/0.977 & 29.660/0.929 & 24.645/0.854 & 27.576/0.906 & 0.321 & 4.98\\
				CAIN~\cite{choi2020channel} & \underline{32.537}/\underline{0.968} & 26.477/0.857 & \underline{39.890}/\underline{0.990} & 35.630/0.978 & 29.998/0.931 & 25.060/0.857 & 28.184/0.911 & 0.071 & 42.78\\
				SoftSplat~\cite{niklaus2020softmax} & 32.835/0.969 & \textcolor{blue}{27.582}/0.881 & \textcolor{blue}{40.165}/\textcolor{blue}{0.991} & \textcolor{blue}{36.017}/\textcolor{blue}{0.979} & 30.604/0.937 & \textcolor{blue}{25.436}/0.864 & 28.813/0.924 & 0.206 & 12.46\\
				EDSC~\cite{cheng2021multiple} & \underline{32.677}/\underline{0.969} & 26.689/0.860 & 39.792/0.990 & 35.283/0.977 & 29.815/0.929 & 24.872/0.854 & 27.641/0.904 & 0.067 & 8.95\\
				XVFI~\cite{sim2021xvfi} & 32.224/0.966 & 26.565/0.863 & 38.849/0.989 & 34.497/0.975 & 29.381/0.929 & 24.677/0.855 & 27.759/0.909 & 0.108 & 5.61\\
				QVI~\cite{xu2019quadratic} & 32.668/0.967 & 27.483/\textcolor{blue}{0.883} & 36.648/0.985 & 34.637/0.978 & \textcolor{blue}{30.614}/\textcolor{blue}{0.947} & 25.426/\textcolor{blue}{0.866} & \textcolor{blue}{28.819}/\textcolor{blue}{0.926} & 0.257 & 29.23\\
				FLAVR~\cite{kalluri2020flavr} & \underline{\textcolor{red}{33.389}}/\underline{\textcolor{red}{0.971}} & 27.450/0.873 & 40.135/0.990 & 35.988/\textcolor{blue}{0.979} & 30.541/0.937 & 25.188/0.860 & 28.487/0.915 & 0.695 & 42.06\\
				\midrule
				ST-MFNet (Ours) & \textcolor{blue}{33.384}/\textcolor{blue}{0.970} & \textcolor{red}{28.287}/\textcolor{red}{0.895} & \textcolor{red}{40.775}/\textcolor{red}{0.992} & \textcolor{red}{37.111}/\textcolor{red}{0.985} & \textcolor{red}{31.698}/\textcolor{red}{0.951} & \textcolor{red}{25.810}/\textcolor{red}{0.874} & \textcolor{red}{29.175}/\textcolor{red}{0.929} & 0.901 & 21.03\\
				\bottomrule
		\end{tabular}}
		\vspace{-2.5mm}
		\caption{Quantitative comparison results (PSNR/SSIM) for ST-MFNet and 14 tested methods. In some cases, underlined scores based on the pre-trained models are provided in the table, when they outperform their re-trained counterparts. OOM denotes cases where our GPU runs out of memory for the evaluation. For each column, the best result is colored in \textcolor{red}{red} and the second best is colored in \textcolor{blue}{blue}. The average runtime (RT) for interpolating a 480p frame as well as the number of model parameters (\#P) for each method are also reported.}
		\label{quantitative}
		\vspace{-6mm}
	\end{center}
\end{table*}

\section{Results and Analysis}\label{sec:exp}
In this section, we analyze our proposed model through ablation studies, and compare it with 14 state-of-the-art methods both quantitatively and qualitatively. 

\subsection{Ablation Study}

The key ablation study results are summarized in Table~\ref{tab:ablation}, where five versions of ST-MFNet have been evaluated. Figure \ref{fig:ablation} provides a visual comparison between the frames generated by each test variant and the full ST-MFNet model. Additional ablation study results, visualizations and analyses are available in Appendix~\ref{sec:abl} and \ref{sec:visualize}.

\noindent\textbf{MIFNet and BLFNet branches.} To verify that the MIFNet and BLFNet branches are both effective, two variants of ST-MFNet, Ours-\textit{w/o MIFNet} and Ours-\textit{w/o BLFNet}, were created by removing the two branches respectively. Both variants were trained and evaluated using the same configurations described above. 
It is observed that, firstly, both Ours-\textit{w/o MIFNet} and Ours-\textit{w/o BLFNet} achieve lower overall performance compared to the full ST-MFNet (Ours). Secondly, compared to Ours, the model performance on large motion (DAVIS) drops more significantly when BLFNet is removed, and that on complex motion (VFITex) degrades more severely when MIFNet is removed. It can also be observed in Figure {\ref{fig:ablation}}, for the case without MIFNet (sub-figures (a-d)), that the model fails to capture the complex motion of the wave. When BLFNet was removed from the original ST-MFNet (sub-figures (e-h)), the occluded region which is also undergoing a large movement has not been interpolated properly. These observations mean that the contribution of each branch aligns well with our original motivation, hence implying the unique advantages of both multi-flow (for complex motion) and single-flow (for large motion) branches have been enabled.

\noindent\textbf{UMSResNext for multi-flow estimation.} To measure the efficacy of the new UMSResNext, we replaced the UMSResNext described in Section~\ref{sec:method_1} with the U-Net used in \cite{lee2020adacof} to predict similar multi-flows. This is denoted as Ours-\textit{unet}. As shown in Table~\ref{tab:ablation}, ST-MFNet with UMSResNext achieves enhanced performance on all test sets, and this is also demonstrated by the visual comparison example in Figure {\ref{fig:ablation}} (i-l). Another advantage of UMSResNext is that it has much fewer parameters ({\raise.17ex\hbox{$\scriptstyle\sim$}}4M) than U-Net ({\raise.17ex\hbox{$\scriptstyle\sim$}}21M).

\noindent\textbf{Texture Enhancement.} The importance of the TENet was also analyzed by training another variant Ours-\textit{w/o TENet}, where the TENet is removed. Table~\ref{tab:ablation} shows that there is a significant performance decrease compared to the full version, especially on DAVIS and VFITex. This demonstrates the contribution of the spatio-temporal filtering on frames over a wider temporal window for content with large and complex motions. Figure~{\ref{fig:ablation}} (m-p) also shows an example, where the full ST-MFNet with the TENet produces richer textural detail compared to the version without TENet.


\noindent\textbf{ST-GAN.} To investigate the effectiveness of the ST-GAN training, we compared the perceptual quality of the interpolated content generated by the fine-tuned network Ours-$\mathcal{L}_{p}$ and the distortion-oriented model Ours-$\mathcal{L}_{lap}$. We also replaced the ST-GAN with two existing GANs used for VFI, FIGAN~\cite{lee2020adacof} and TGAN~\cite{saito2017temporal}. Example frames produced by these variants are shown in Figure~{\ref{fig:ablation}} (q-v), where the result generated by Ours-$\mathcal{L}_{p}$ exhibits sharper edges and clearer structures compared to those produced by other variants. Quantitative evaluation results of these variants based on LPIPS are provided in Appendix~\ref{sec:abl}.

\begin{figure*}[t]
    \centering
    \subfloat {\includegraphics[width=0.107\linewidth]{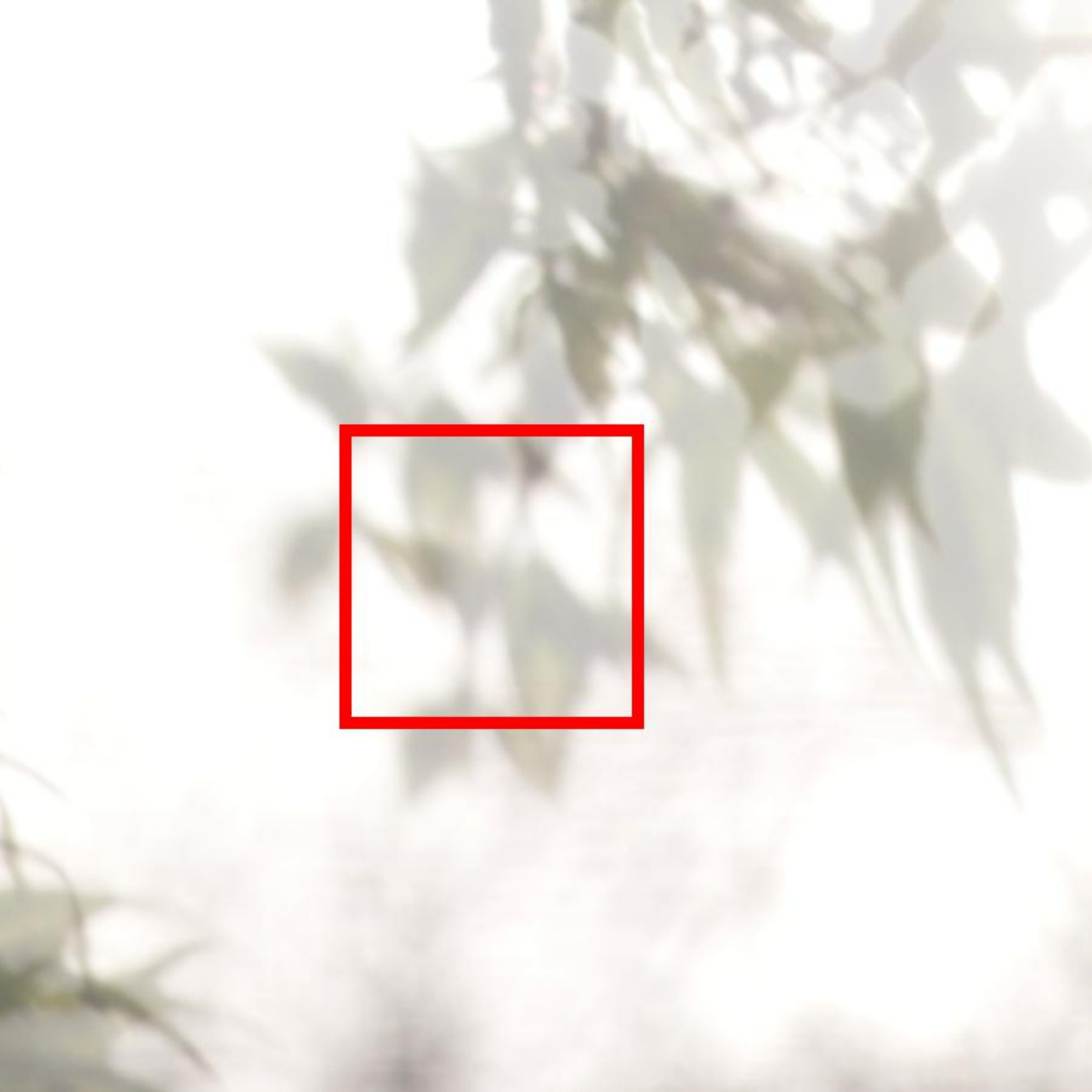}}\;\!\!
	\subfloat {\includegraphics[width=0.107\linewidth]{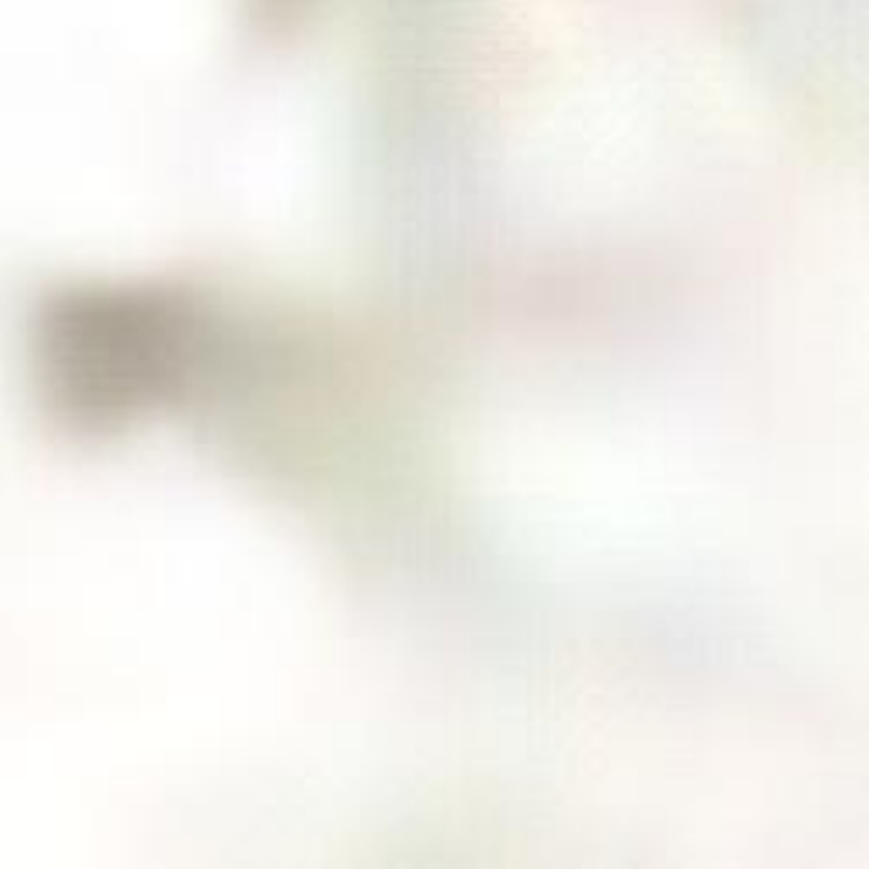}}\;\!\!
	\subfloat {\includegraphics[width=0.107\linewidth]{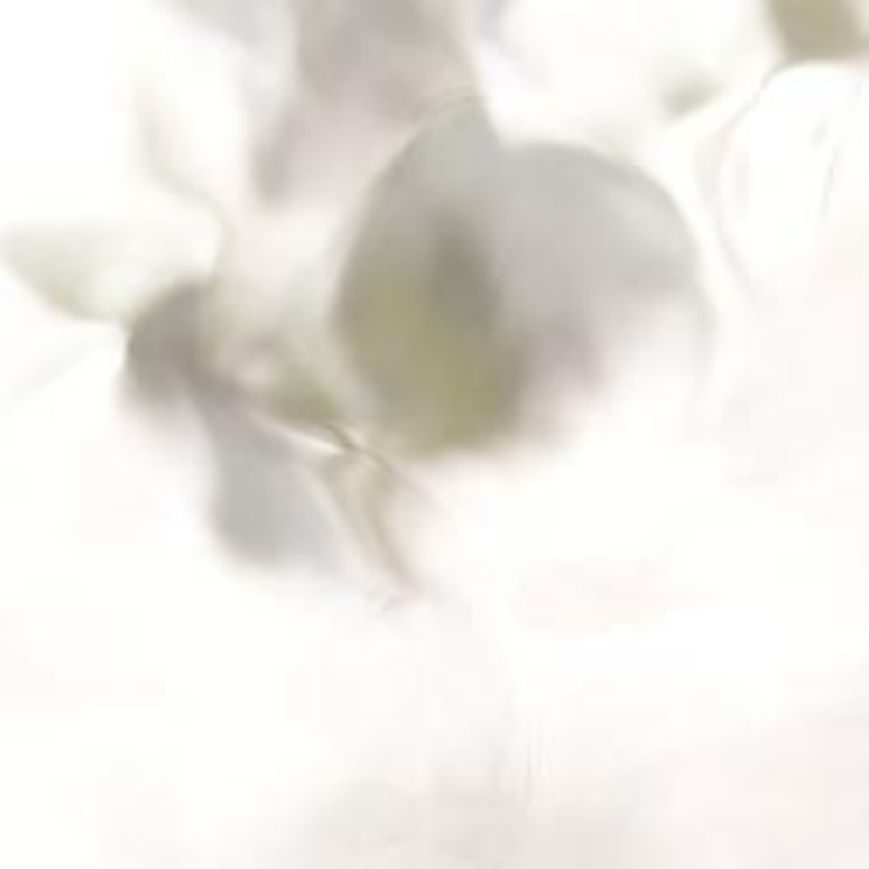}}\;\!\!
    \subfloat {\includegraphics[width=0.107\linewidth]{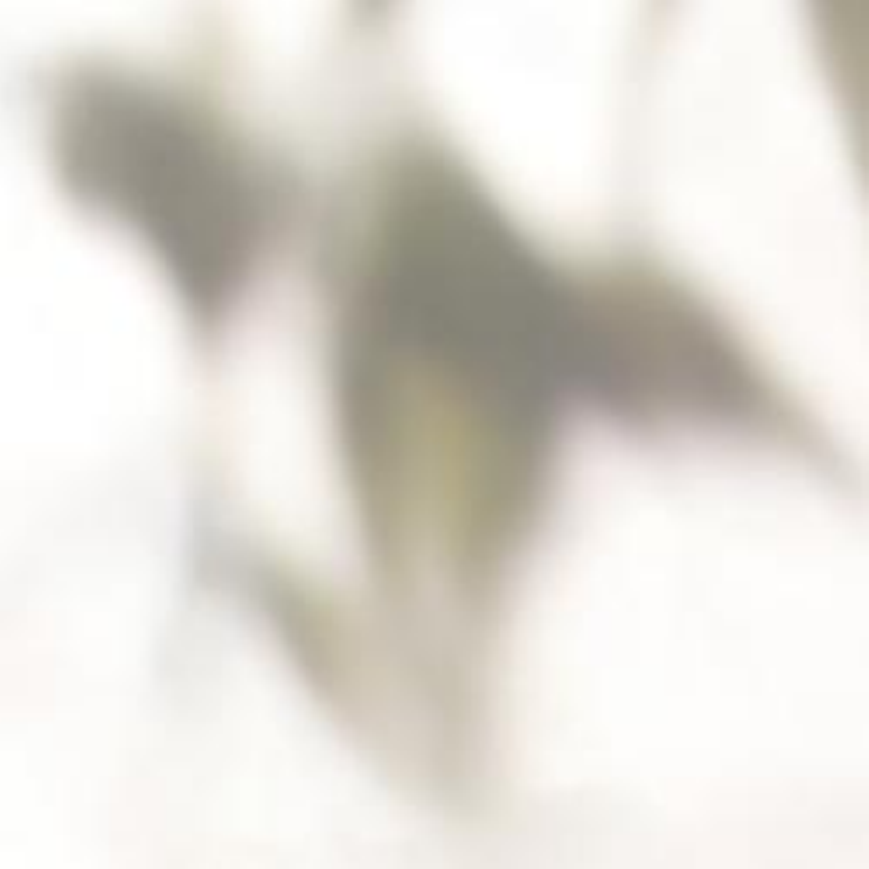}}\;\!\!
    \subfloat {\includegraphics[width=0.107\linewidth]{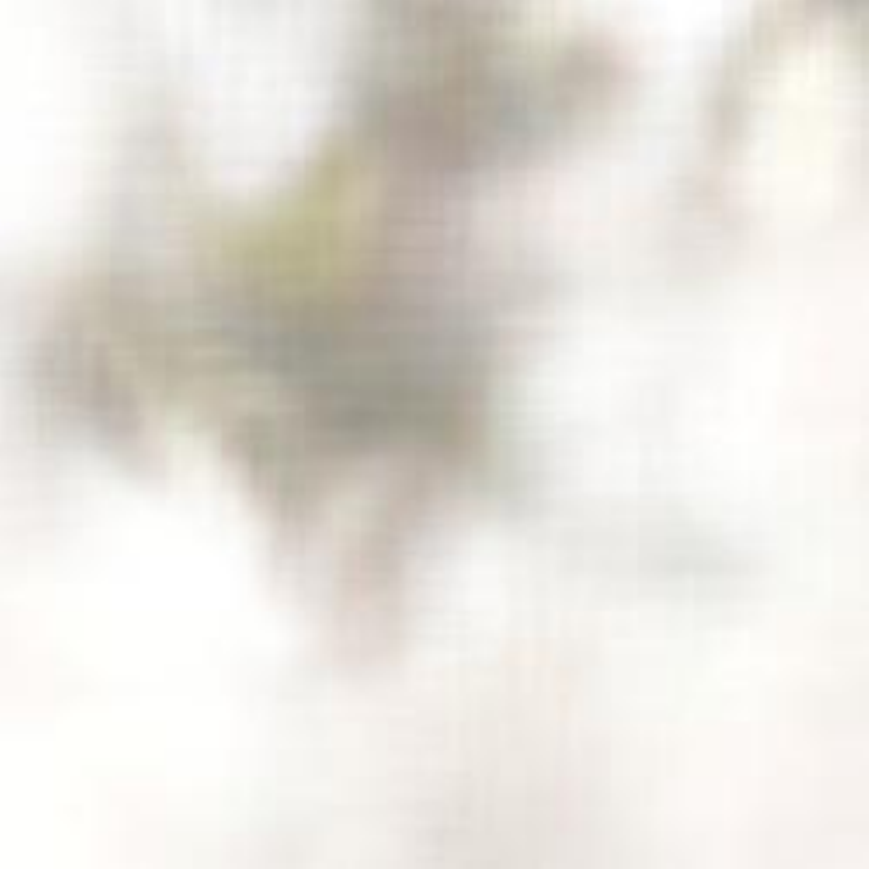}}\;\!\!
    \subfloat {\includegraphics[width=0.107\linewidth]{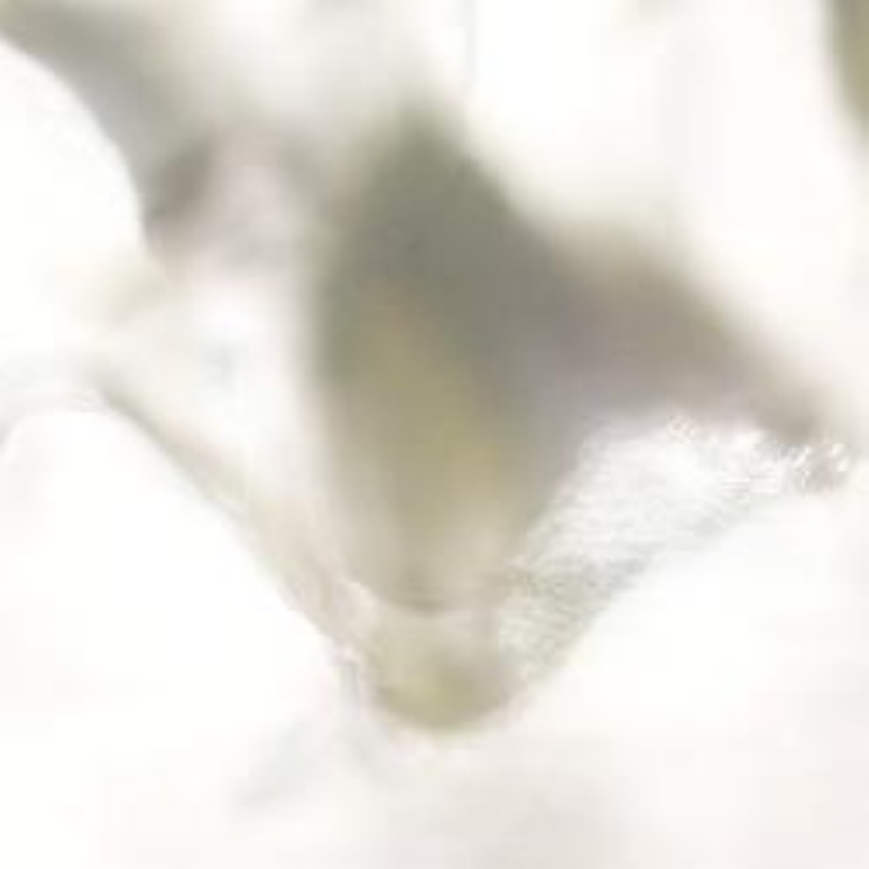}}\;\!\!
    \subfloat {\includegraphics[width=0.107\linewidth]{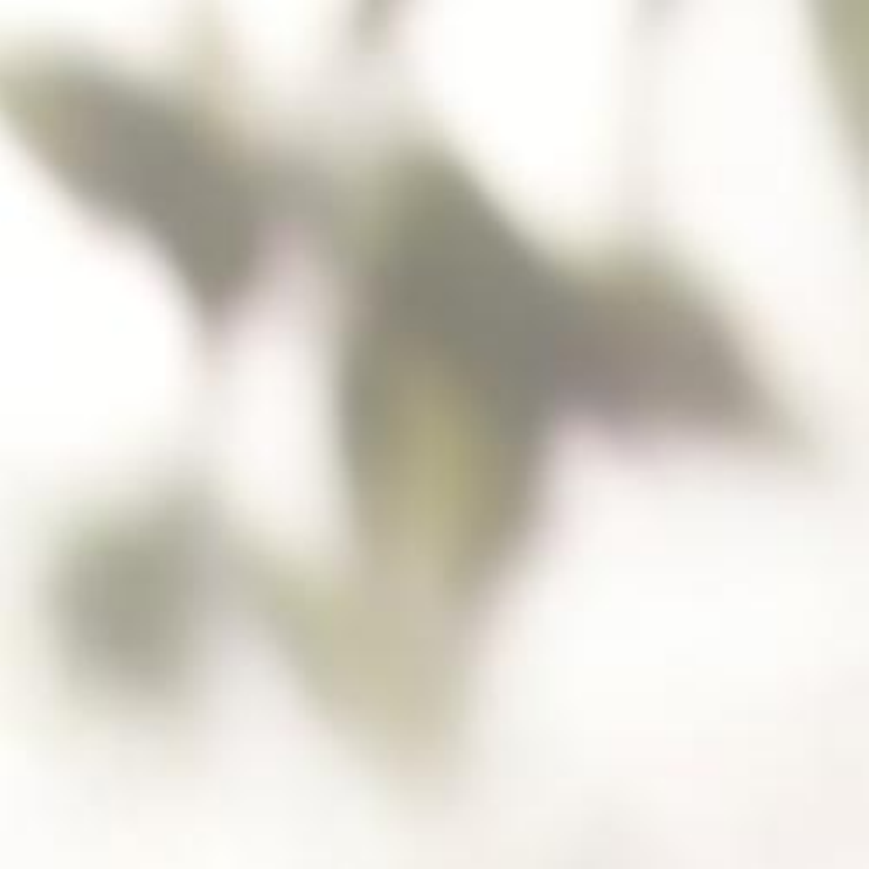}}\;\!\!
    \subfloat {\includegraphics[width=0.107\linewidth]{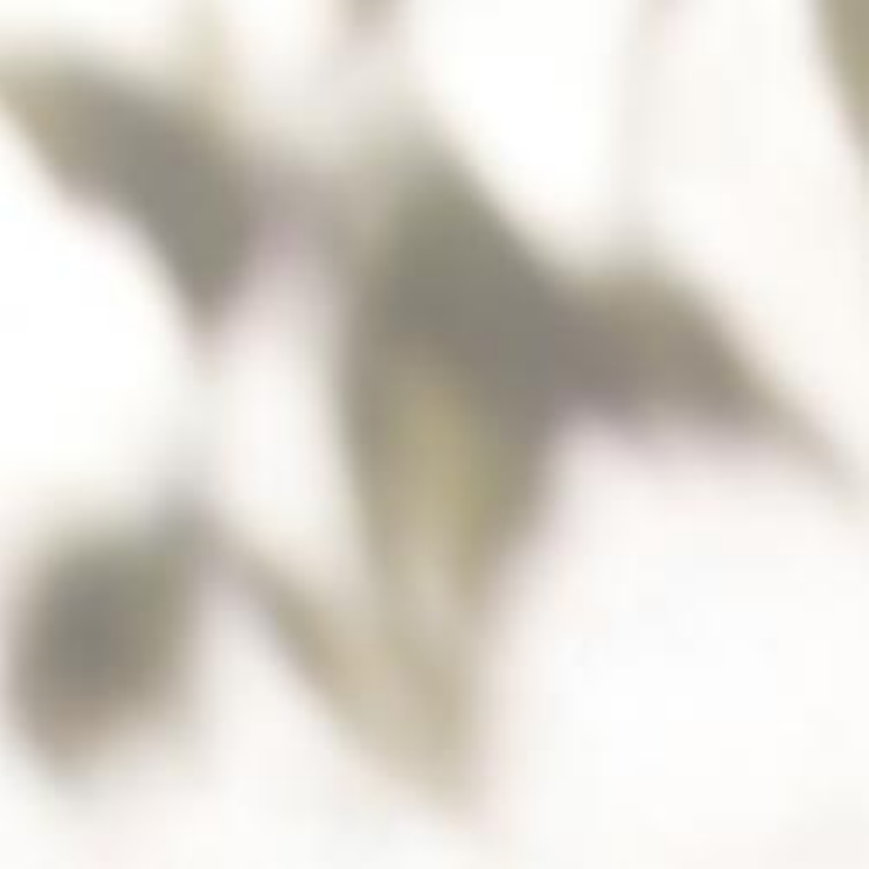}}\;\!\!
    \subfloat {\includegraphics[width=0.107\linewidth]{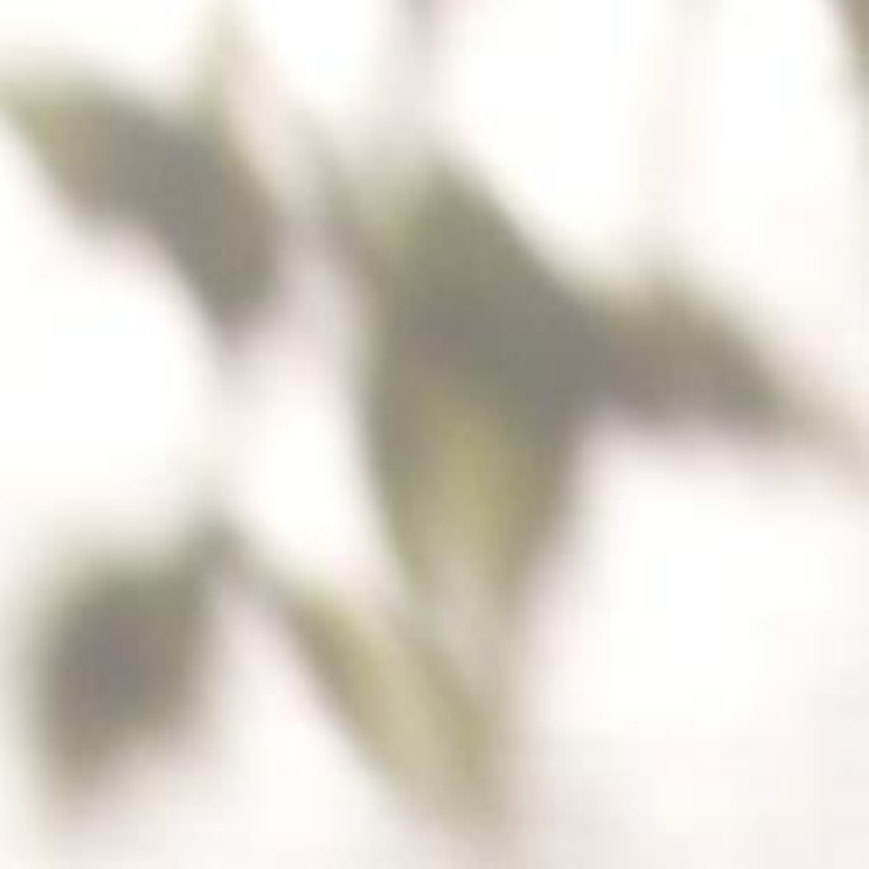}}\\

    \subfloat {\includegraphics[width=0.107\linewidth]{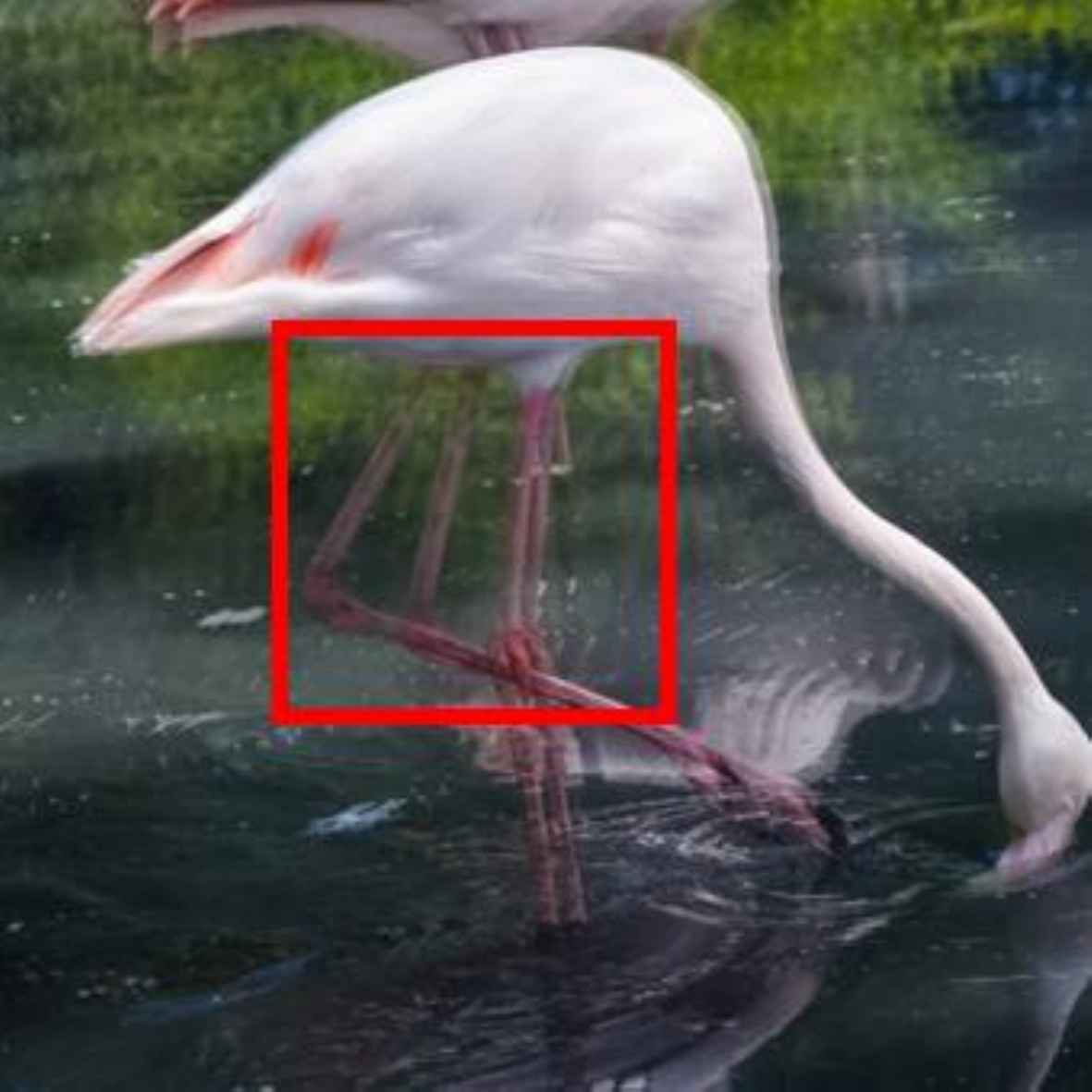}}\;\!\!
	\subfloat {\includegraphics[width=0.107\linewidth]{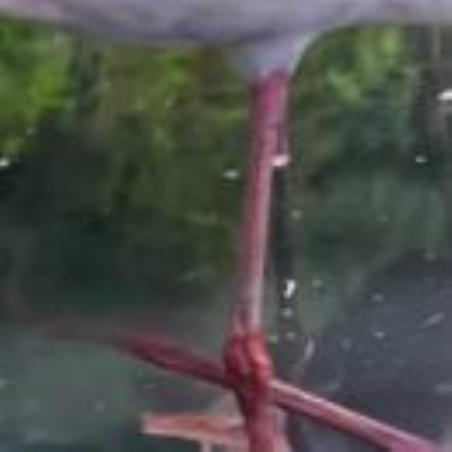}}\;\!\!
	\subfloat {\includegraphics[width=0.107\linewidth]{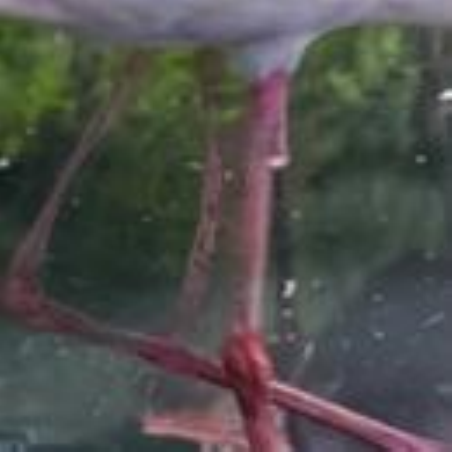}}\;\!\!
    \subfloat {\includegraphics[width=0.107\linewidth]{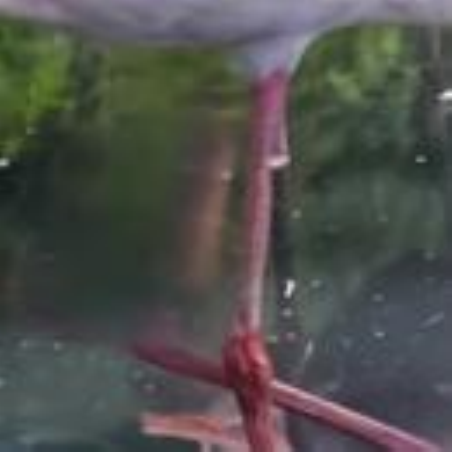}}\;\!\!
    \subfloat {\includegraphics[width=0.107\linewidth]{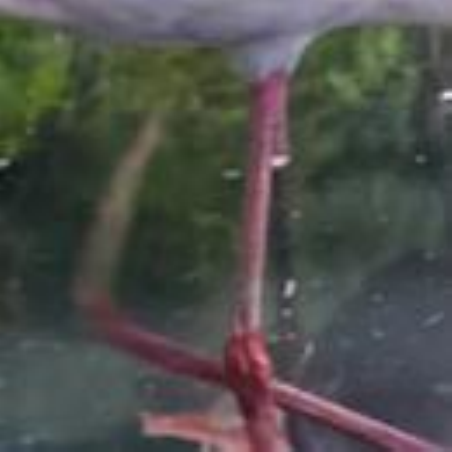}}\;\!\!
    \subfloat {\includegraphics[width=0.107\linewidth]{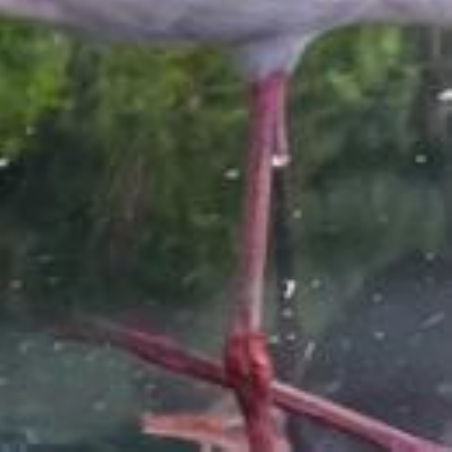}}\;\!\!
    \subfloat {\includegraphics[width=0.107\linewidth]{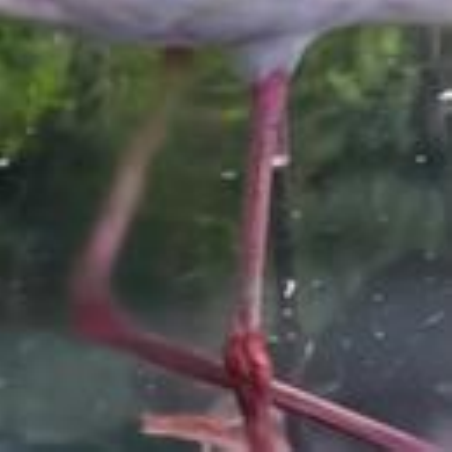}}\;\!\!
    \subfloat {\includegraphics[width=0.107\linewidth]{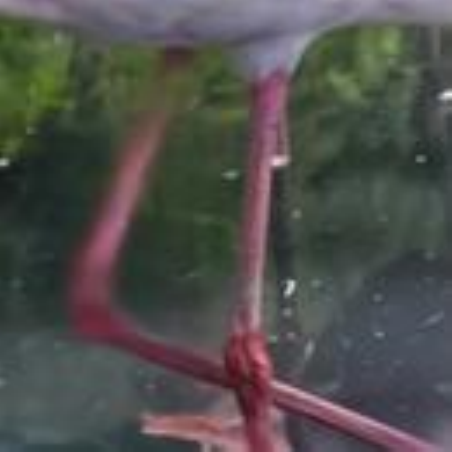}}\;\!\!
    \subfloat {\includegraphics[width=0.107\linewidth]{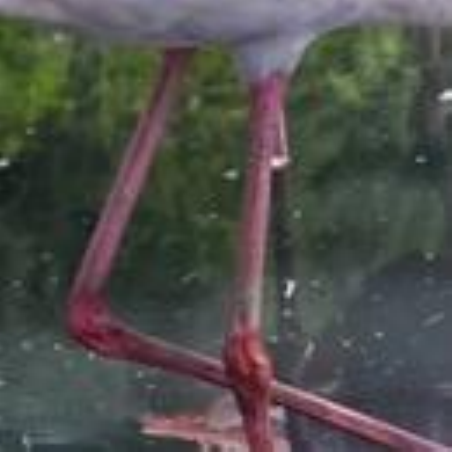}}\\

    \setcounter{subfigure}{0}
    \subfloat[Overlay] {\includegraphics[width=0.107\linewidth]{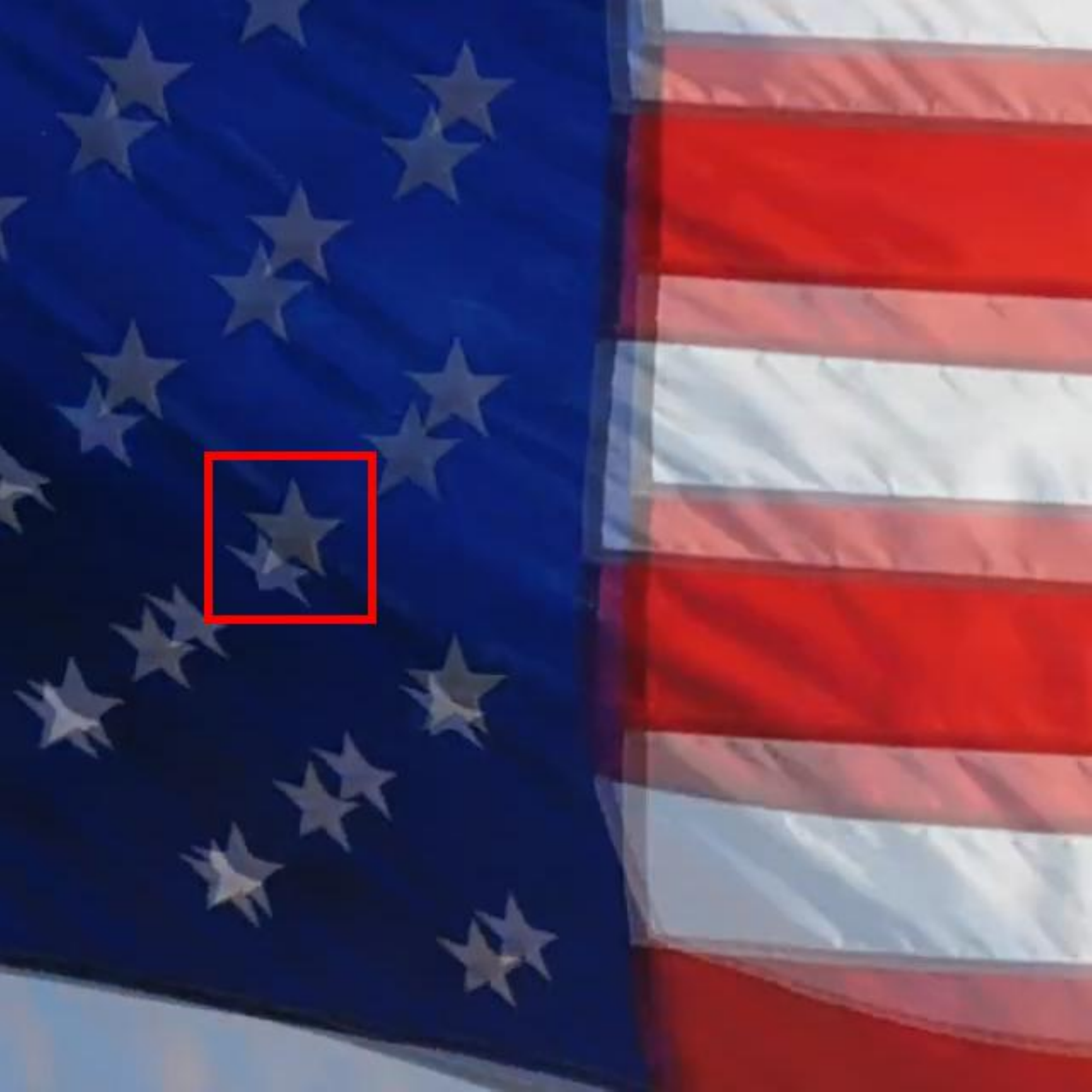}}\;\!\!
	\subfloat[BMBC] {\includegraphics[width=0.107\linewidth]{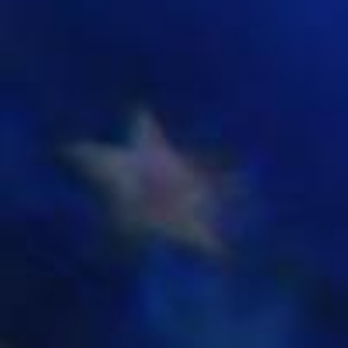}}\;\!\!
	\subfloat[DAIN] {\includegraphics[width=0.107\linewidth]{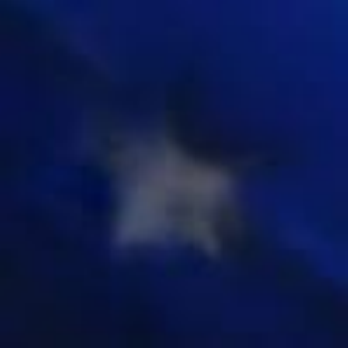}}\;\!\!
    \subfloat[Softsplat] {\includegraphics[width=0.107\linewidth]{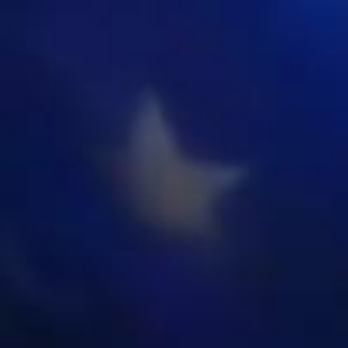}}\;\!\!
    \subfloat[FLAVR] {\includegraphics[width=0.107\linewidth]{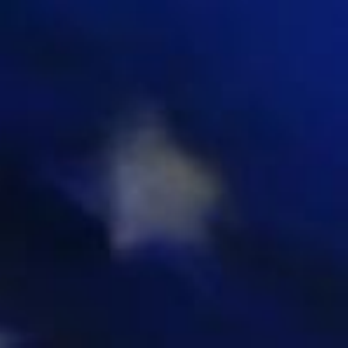}}\;\!\!
    \subfloat[QVI] {\includegraphics[width=0.107\linewidth]{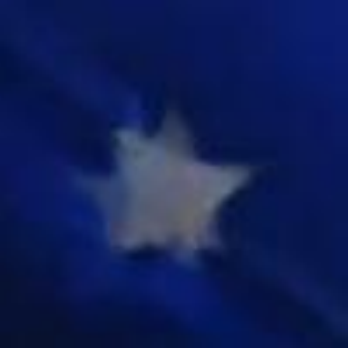}}\;\!\!
    \subfloat[Ours] {\includegraphics[width=0.107\linewidth]{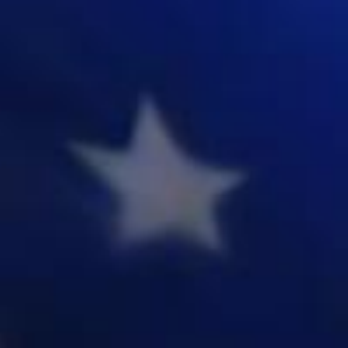}}\;\!\!
    \subfloat[Ours-$\mathcal{L}_p$] {\includegraphics[width=0.107\linewidth]{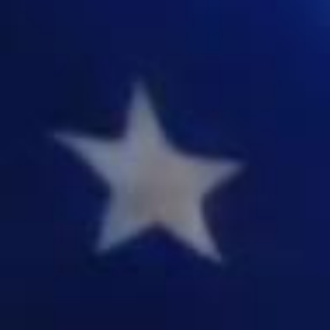}}\;\!\!
    \subfloat[GT] {\includegraphics[width=0.107\linewidth]{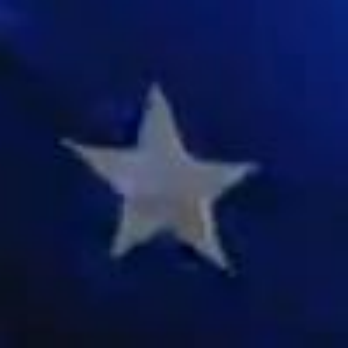}}\\[-0.9em]
    
    \vspace*{0.05cm}
    \caption{Qualitative interpolation examples by different methods. The first column (a) shows the overlaid adjacent frames. Columns (b-f) correspond to some of the best-performing benchmark methods. The results of our distortion-oriented model (g) and perception-oriented model (h) are also included, along with the ground truth frames (i). Video comparison examples can be found in Appendix~\ref{sec:video}.}
	\label{fig:qualitative}
    \vspace*{-0.4cm}
\end{figure*}


\begin{figure}[t]
\begin{center}
   \includegraphics[width=0.9\linewidth]{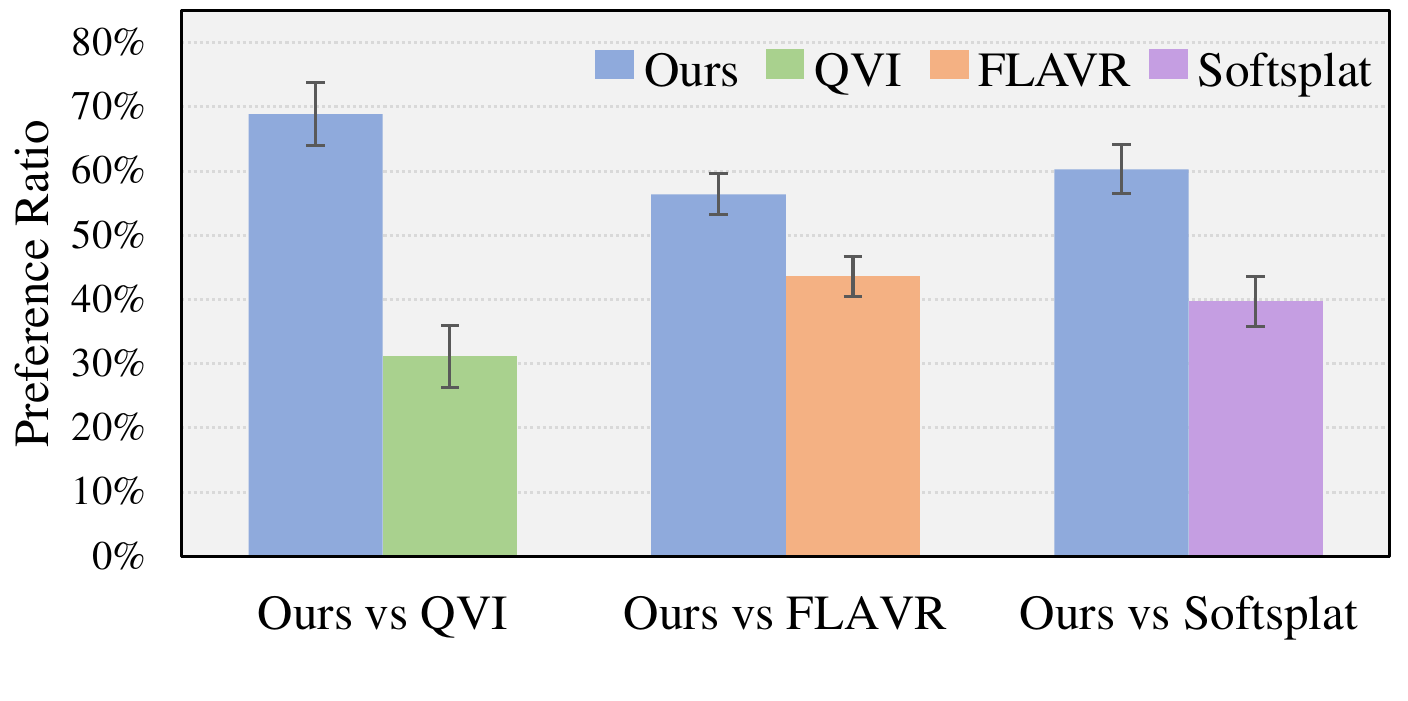}
\end{center}
\vspace{-9mm}
\caption{\label{fig:userstudy}
Results of the user study showing preference ratios for the tested interpolation methods. The error bars denote standard deviation over test videos.}
\vspace{-6mm}
\end{figure}

\subsection{Quantitative Evaluation}

We compared the proposed ST-MFNet with 14 state-of-the-art VFI models including DVF~\cite{liu2017video}, SuperSloMo~\cite{jiang2018super}, SepConv~\cite{niklaus2017video}, DAIN~\cite{bao2019depth}, BMBC~\cite{park2020bmbc}, AdaCoF~\cite{lee2020adacof}, FeFlow~\cite{gui2020featureflow}, CDFI~\cite{ding2021cdfi}, CAIN~\cite{choi2020channel}, Softsplat~\cite{niklaus2020softmax}, EDSC~\cite{cheng2021multiple}, XVFI~\cite{sim2021xvfi}, QVI~\cite{xu2019quadratic} and FLAVR~\cite{kalluri2020flavr}. For fair comparison, we re-trained all benchmark models with the same training and validation datasets used for ST-MFNet under identical training configurations. The comprehensive evaluation results are summarized in Table~\ref{quantitative}. For all these models, we additionally evaluated their pre-trained versions provided in the original literature (where applicable). If the pre-trained results are better than the re-trained counterparts, the former are presented and underlined.

Two key observations can be made from Table~\ref{quantitative}. Firstly, by using our training set (Vimeo-90k+BVI-DVC), the re-trained performance of all compared models has been improved over their pre-trained versions on large and complex motions, i.e. on DAVIS, SNU-FILM (medium, hard, extreme) and VFITex. For seven models, the pre-trained versions achieved higher PSNR and SSIM values on the UCF-101 dataset. This may be due to the similar characteristics between their pre-training dataset, Vimeo-90k and UCF-101. We also noted that our ST-MFNet offers the best results for DAVIS, SNU-FILM (all subsets) and VFITex, with a significant improvement of 0.36-1.09dB (PSNR) over the runner-up for each test set. It is only outperformed by the pre-trained FLAVR on UCF101 with marginal difference of 0.005dB (PSNR) and 0.001 (SSIM). This demonstrates the excellent generalization ability of the proposed ST-MFNet. Additional evaluation results in terms of LPIPS, and results on 4$\times$/8$\times$ interpolation can be found in Appendix~\ref{sec:4x8x}.

\noindent\textbf{Complexity.} The model complexity was measured on the 480p sequences from DAVIS test set. The average runtime (RT, in seconds) for interpolating one frame is reported in Table~\ref{quantitative} for each tested network, alongside its total number of parameters. We noticed that ST-MFNet has a relatively high computational complexity among all tested models. The reduction of model complexity remains one of our future works.

\subsection{Qualitative Evaluation}\label{sec:qualitative}

\noindent\textbf{Visual comparisons.}
Examples frames interpolated by our model and several best-performing state-of-the-art methods are shown in Figure~\ref{fig:qualitative}. It can be observed that the results generated by the perceptually trained ST-MFNet (Ours-$\mathcal{L}_p$) are closer to the ground truth, containing fewer visual artifacts and exhibiting better perceptual quality.

\noindent\textbf{User Study.}
As single frames cannot fully reflect the perceptual quality of interpolated content, we conducted a user study where our method was compared against three competitive benchmark approaches, QVI, FLAVR and Softsplat (re-trained using its original perceptual loss~\cite{niklaus2020softmax}). For this study, 20 videos randomly selected from DAVIS, SNU-FILM and VFITex were used as the test content for the three tested models. In each trial of a test session, participants were shown a pair of videos including one interpolated by perceptually optimized ST-MFNet and the other one by QVI, FLAVR or Softsplat. This results in a total of 60 trials in each test session. The order of video presentation was randomized in each trial (the order of trials was also random), and the subject in each case was asked to choose the sequence with higher perceived quality. Twenty subjects were paid to participate in this study. See more details of the user study in Appendix~\ref{sec:study}. 

The collected user study results are summarized in Figure~\ref{fig:userstudy}. We observed that approaching 70\% of users on average preferred ST-MFNet against QVI, and this figure is statistically significant for 95\% confidence based on a t-test experiment ($p<.00000003$). The average preference difference between our method and FLAVR is smaller, with 56\% users in favor of ST-MFNet results. This was also significant at a 95\% confidence level ($p<.0001$). Finally, when comparing against Softsplat, around 60\% of subjects favored our method, where the significance holds again at 95\% level ($p<.000005$).

\section{Limitations and Potential Negative Impacts}
Although superior interpolation performance has been observed from the proposed method, we are aware of the relatively low inference speed associated with this model. This is mainly due to its large network capacity. Training such large models can also potentially introduce negative impact on the environment due to the significant power consumption of computational hardware~\cite{lacoste2019quantifying}. This can be mitigated through model complexity reduction based on network compression~\cite{modelcompression} and knowledge distillation~\cite{hinton2015distilling}.

\section{Conclusion}

In this paper, we propose a novel video frame interpolation algorithm, ST-MFNet, which consistently achieved improved interpolation performance (up to a 1.09dB PSNR gain) over state-of-the-art methods on various challenging video content. The proposed method features three main innovative design elements. Firstly, flexible many-to-one multi-flows were combined with conventional one-to-one optical flows in a multi-branch fashion, which enhances the ability of capturing large and complex motions. Secondly, a novel architecture was designed to predict multi-interflows at multiple scales, leading to reduced complexity but enhanced performance. Thirdly, we employed a 3D CNN architecture and the ST-GAN originally proposed for texture synthesis to enhance the visual quality of textures in the interpolated content. Our quantitative and qualitative experiments showed that all of these contribute to the final performance of our model, which consistently outperforms many state-of-the-art methods with significant gains.
\\
\\
\noindent\textbf{Acknowledgment.} This work was funded by the China Scholarship Council, University of Bristol, and the UKRI MyWorld Strength in Places Programme (SIPF00006/1).

\appendix
\section*{Appendix}

\section{Discriminator for ST-MFNet}
\label{sec:dis}

\begin{figure*}[htbp]
\begin{center}
   \includegraphics[width=0.85\linewidth]{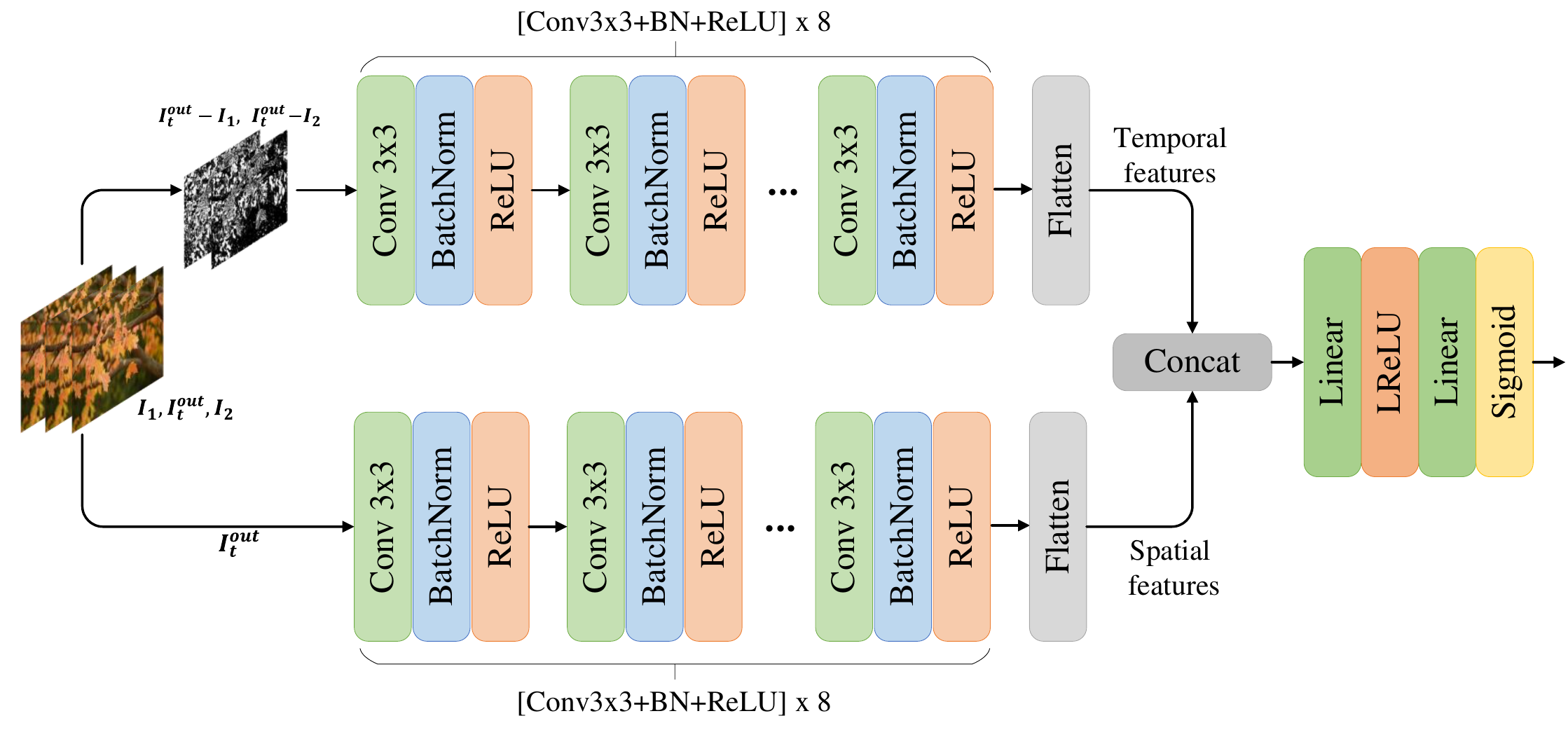}
\end{center}
\vspace{-6mm}
\caption{\label{fig:d}
Architecture of the discriminator used for training ST-MFNet.}
\end{figure*}

The architecture of the discriminator employed in this work is illustrated in Figure~\ref{fig:d}; this was originally designed to train ST-GAN~\cite{yang2021spatiotemporal} for texture synthesis. It contains a temporal and a spatial branch. The former takes the differences between the interpolated output $I_t^{out}$ (where $t=1.5$) of ST-MFNet and its two adjacent original frames $I_1,I_2$ as input. The differences here represent the high-frequency temporal information within these three frames. The spatial branch in this network processes the ST-MFNet output $I_t^{out}$ to generate spatial features. Finally, the temporal and spatial features generated in these two branches are concatenated before fed into the final fully connected layers. 

\section{Additional Ablation Study Results}
\label{sec:abl}

In the main paper, we presented key ablation study results where the primary contributions in the proposed ST-MFNet are evaluated. Here the effectiveness of the up-sampling scale is further investigated, which has been employed during the multi-flow prediction in the MIFNet branch (see Section 3.1 of the main paper). In addition, we present the quantitative ablation study results for the ST-GAN in terms of a perceptually-oriented metric, the Learned Perceptual Image Patch Similarity (LPIPS)~\cite{zhang2018unreasonable}.

\noindent\textbf{Up-sampling}. To evaluate the contribution of the up-sampling scale during the multi-flow prediction, the version of ST-MFNet (Ours-\textit{w/o US}) with only two multi-flow estimation heads (at $l=0,1$ scales) were implemented. It was also trained and evaluated using the same configurations described in the main paper. Its interpolation results are summarized in Table \ref{tab:ablation} alongside more comprehensive ablation study results for the other four variants of ST-MFNet (described in the main paper). It can be observed that Ours-\textit{w/o US} was outperformed by the full version of ST-MFNet (Ours) on all test datasets. The performance difference can also be demonstrated through visual comparison as shown in Figure~\ref{fig:ablation}. All of these confirm the effectiveness of the up-sampling scale in multi-flow estimation. 
\begin{table*}[htbp]
	\begin{center}
		\resizebox{\linewidth}{!}{
			\begin{tabular}{lccccccc}
				\toprule
				& \multirow{2}[1]{*}{UCF101} & \multirow{2}[1]{*}{DAVIS} & \multicolumn{4}{c}{SNU-FILM} & \multirow{2}[1]{*}{VFITex}  \\
				\cmidrule(l{5pt}r{5pt}){4-7}
				& & &Easy&Medium&Hard&Extreme \\
				\midrule
				Ours-\textit{w/o BLFNet} & 33.218/0.970 & 27.767/0.881 & 40.655/0.990 & 36.890/0.984 & 31.205/0.947 & 25.492/0.869 & 28.498/0.915\\
				Ours-\textit{w/o MIFNet} & 33.202/0.969 & 27.886/0.889 & 40.331/0.991 & 36.530/0.982 & 31.321/0.949 & 25.620/0.871 & 28.357/0.911 \\
				Ours-\textit{w/o TENet} & 32.895/0.970 & 27.484/0.880 & 40.275/0.991 & 35.983/0.980 & 30.527/0.937 & 25.374/0.864 & 28.241/0.910 \\ 
				Ours-\textit{unet} & 33.378/0.970 & 28.096/0.892 & 40.616/0.991 & 36.797/0.984 & 31.383/0.950 & 25.680/.
				872& 28.898/0.925 \\
				Ours-\textit{w/o US} & 33.371/0.970 & 28.155/0.893 & 40.248/0.990 & 36.689/0.983 & 31.384/0.949 & 25.636/0.873 & 28.977/0.925 \\
				\midrule
				Ours & 33.384/0.970 & 28.287/0.895 & 40.775/0.992 & 37.111/0.985 & 31.698/0.951 & 25.810/0.874 & 29.175/0.929\\
				\bottomrule
		\end{tabular}}
		\vspace{-1mm}
		\caption{Comprehensive ablation study results on ST-MFNet.}
		\label{tab:ablation}
		\vspace{-3mm}
	\end{center}
\end{table*}

\begin{table*}[h!]
	\begin{center}
			\begin{tabular}{lccccccc}
				\toprule
				& \multirow{2}[1]{*}{UCF101} & \multirow{2}[1]{*}{DAVIS} & \multicolumn{4}{c}{SNU-FILM} & \multirow{2}[1]{*}{VFITex}  \\
				\cmidrule(l{5pt}r{5pt}){4-7}
				& & &Easy&Medium&Hard&Extreme \\
				\midrule
				Ours-$\mathcal{L}_{lap}$ & 0.036 & 0.125 & 0.019 & 0.036 & 0.073 & 0.148 & 0.216\\
				TGAN & 0.034 & 0.117 & 0.019 & 0.033 & 0.068 & 0.142 & 0.213\\
				FIGAN & 0.036 & 0.119 & 0.020 & 0.035 & 0.070 & 0.146 & 0.216\\ 
				Ours-$\mathcal{L}_p$ & 0.033 & 0.116 & 0.017 & 0.031 & 0.065 & 0.140 & 0.210 \\
				\bottomrule
		\end{tabular}
		\vspace{-1mm}
		\caption{Quantitative ablation study results for ST-GAN, in terms of LPIPS.}
		\label{tab:stgan}
		\vspace{-3mm}
	\end{center}
	
\end{table*}
\begin{figure*}[h!]
	\begin{center}
		\subfloat[Overlay]
		{\includegraphics[width=0.234\linewidth]{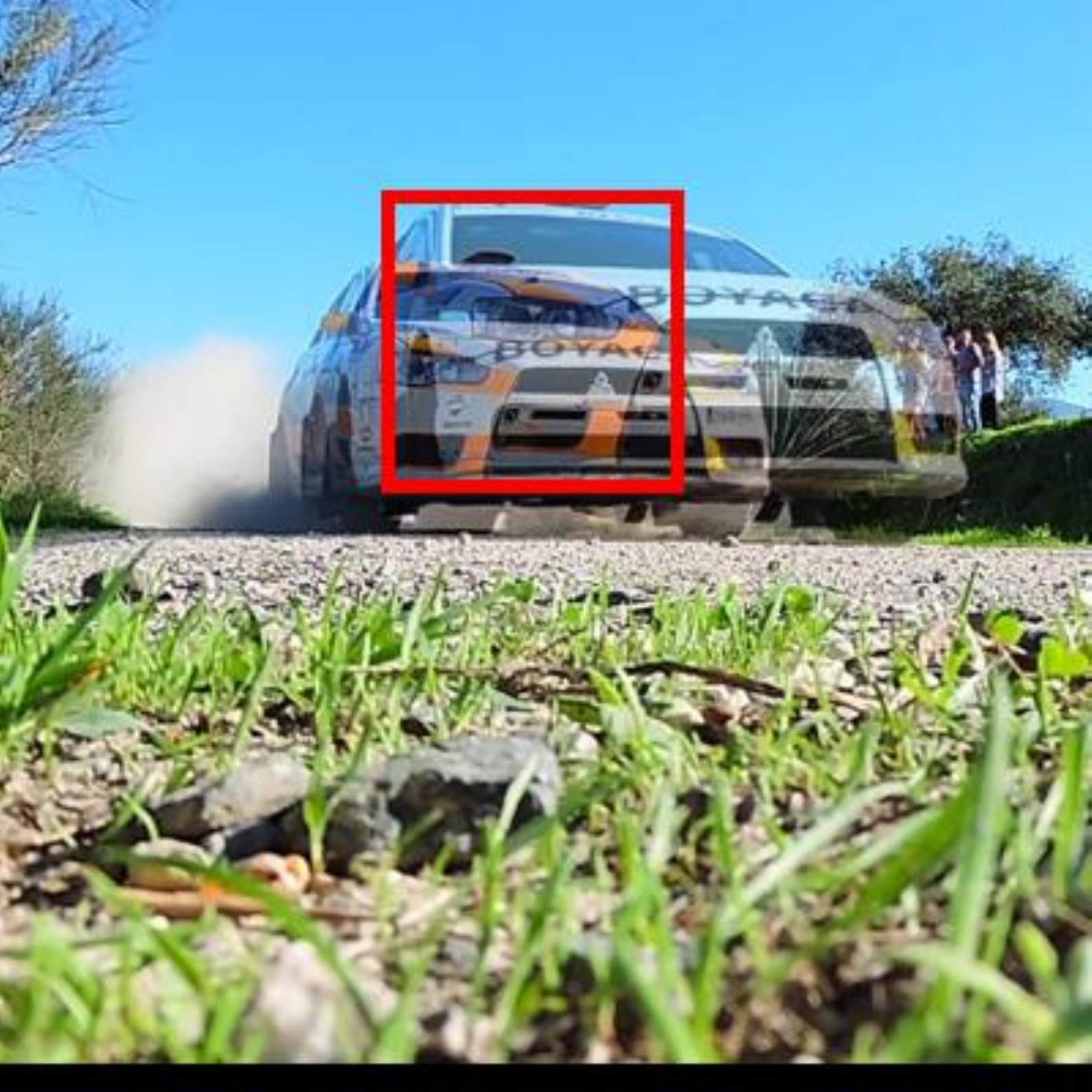}}\,
		\subfloat[GT]
		{\includegraphics[width=0.234\linewidth]{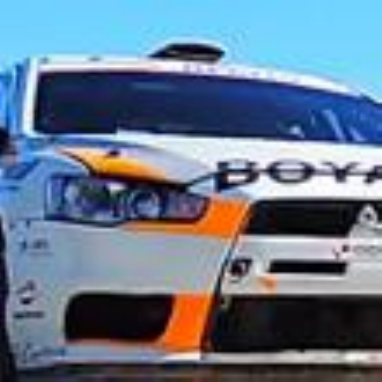}}\,
		\subfloat[Ours-\textit{w/o US}]
		{\includegraphics[width=0.234\linewidth]{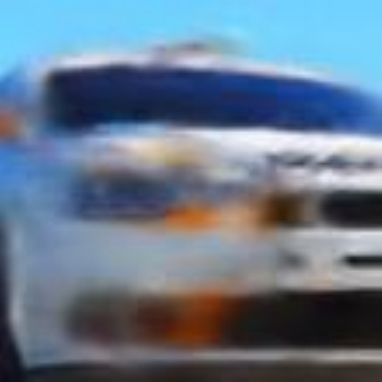}}\,
		\subfloat[Ours-\textit{w/ US}]
		{\includegraphics[width=0.234\linewidth]{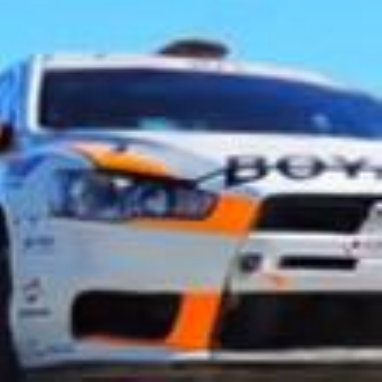}}
 \end{center}
\vspace{-5mm}
\caption{Qualitative results interpolated by the ST-MFNet with the up-sampled scale removed (Ours-\textit{w/o US}) and the full version of ST-MFNet (Ours-\textit{w/ US}). Here ``Overlay" means the overlaid adjacent frames.}
\label{fig:ablation}
\vspace{-3mm}
\end{figure*}

\begin{figure*}[t]
    \centering
    \subfloat {\includegraphics[width=0.138\linewidth]{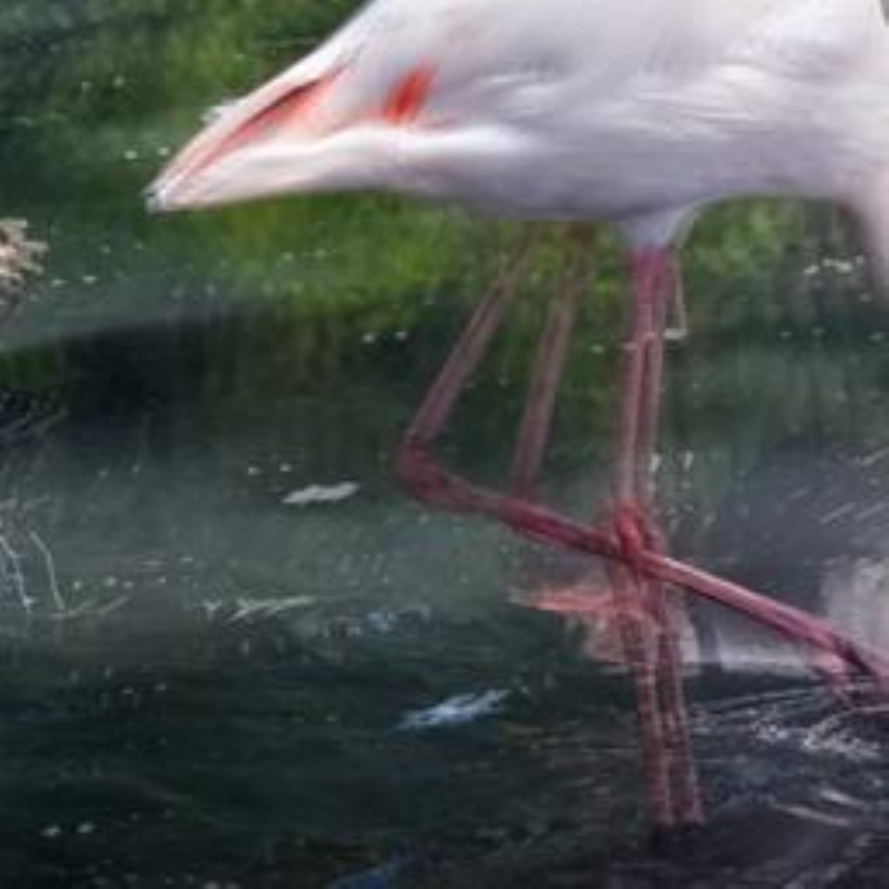}}\;\!\!
	\subfloat {\includegraphics[width=0.138\linewidth]{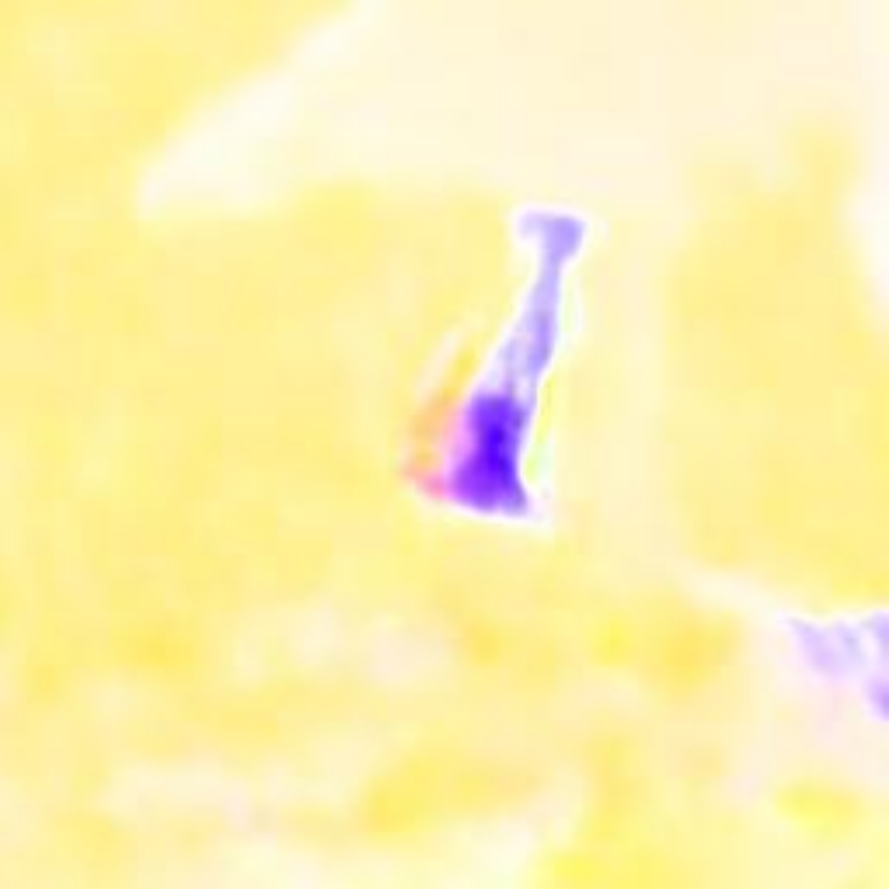}}\;\!\!
	\subfloat {\includegraphics[width=0.138\linewidth]{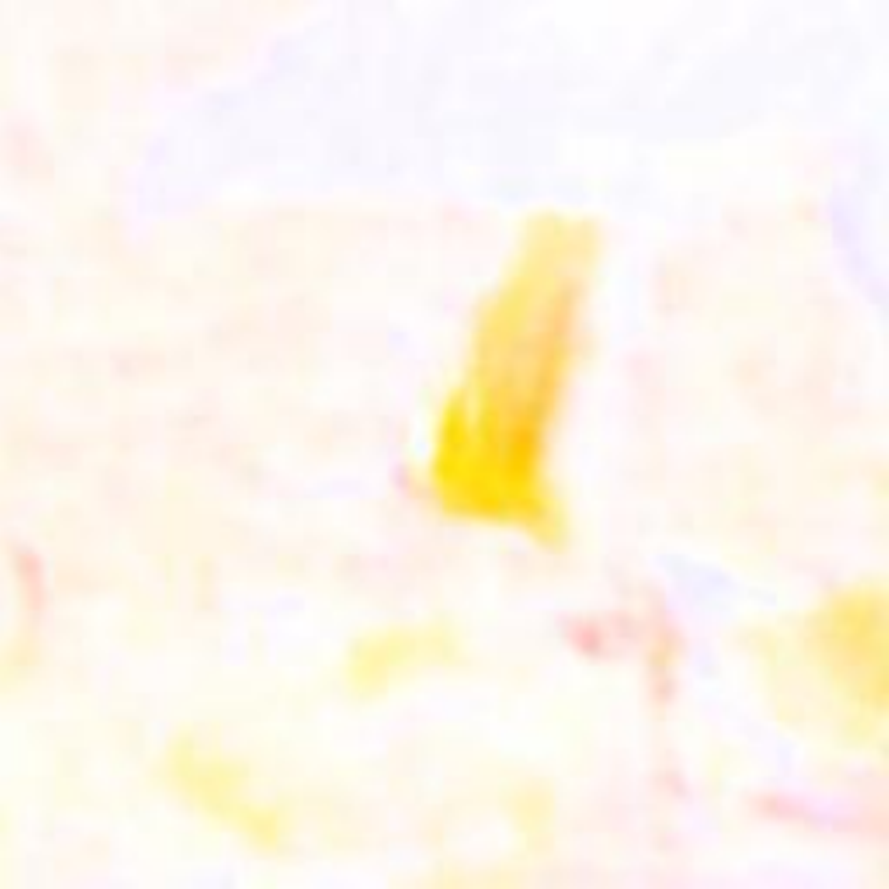}}\;\!\!
    \subfloat {\includegraphics[width=0.138\linewidth]{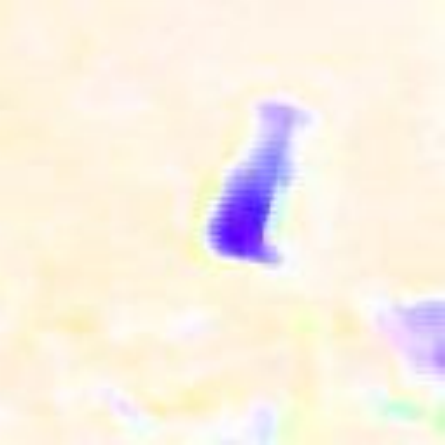}}\;\!\!
    \subfloat {\includegraphics[width=0.138\linewidth]{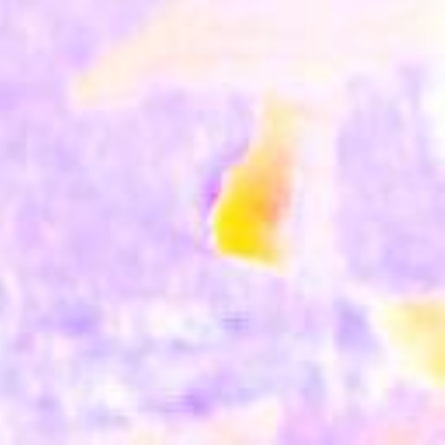}}\;\!\!
    \subfloat {\includegraphics[width=0.138\linewidth]{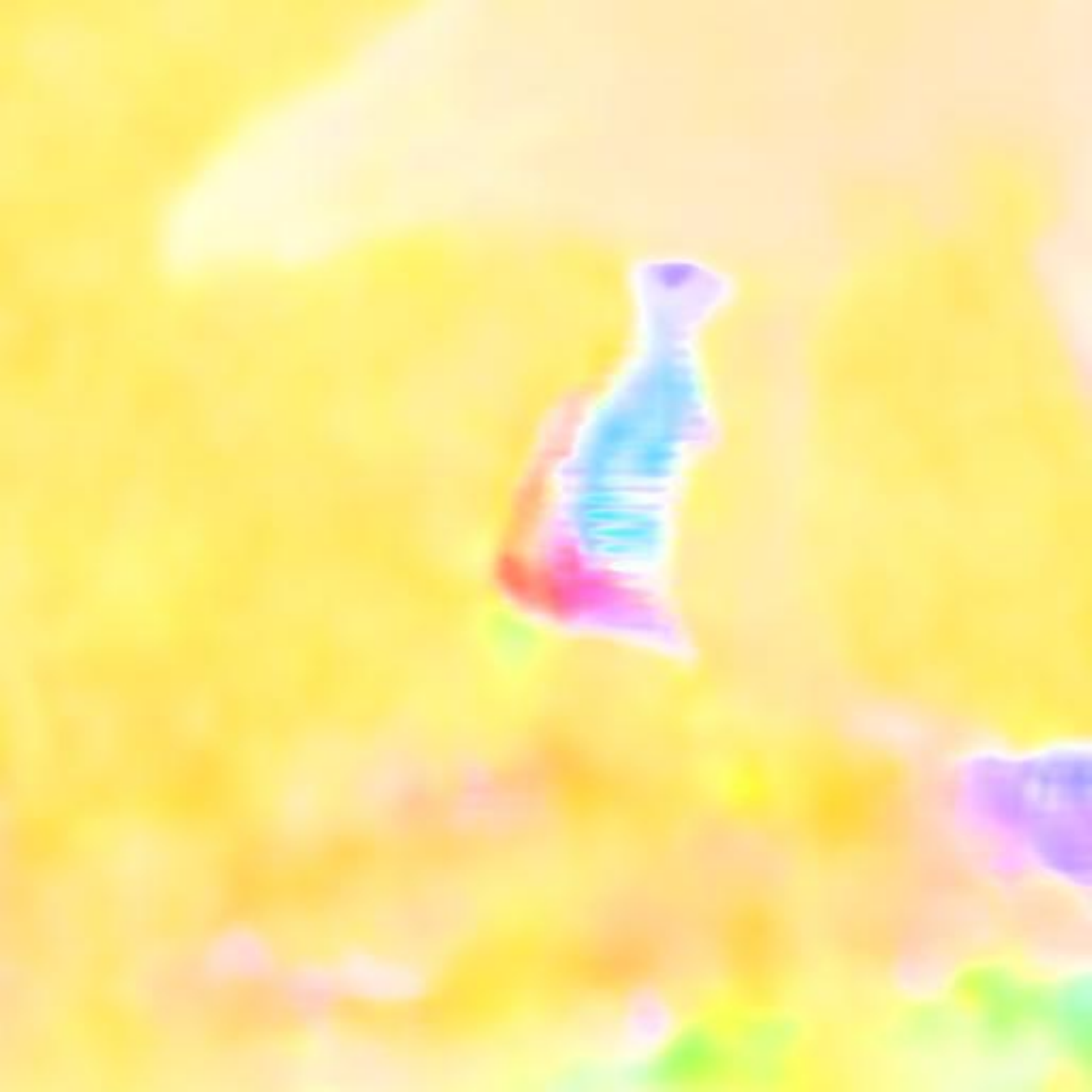}}\;\!\!
    \subfloat {\includegraphics[width=0.138\linewidth]{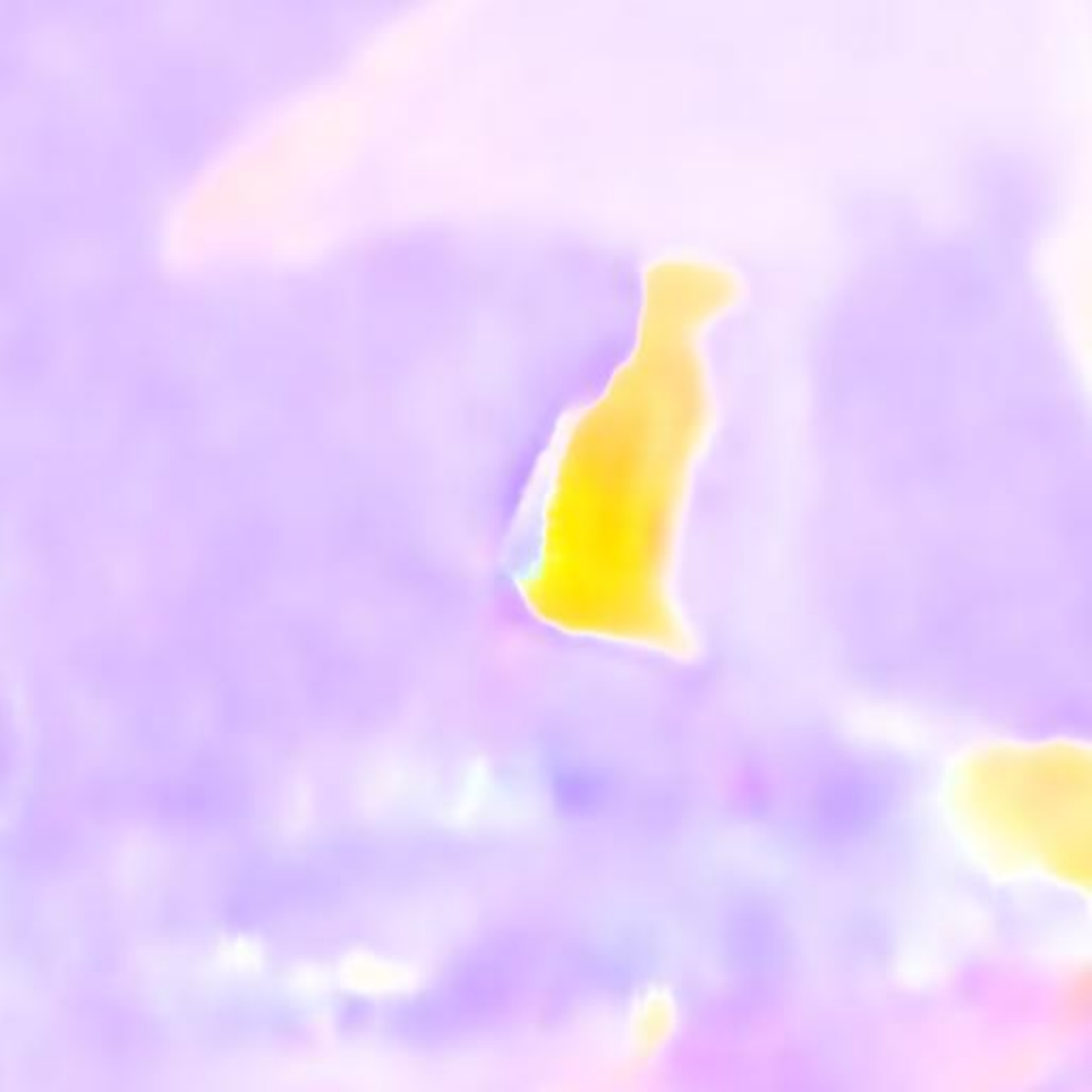}}\\
    \setcounter{subfigure}{0}
    \subfloat[Overlay] {\includegraphics[width=0.138\linewidth]{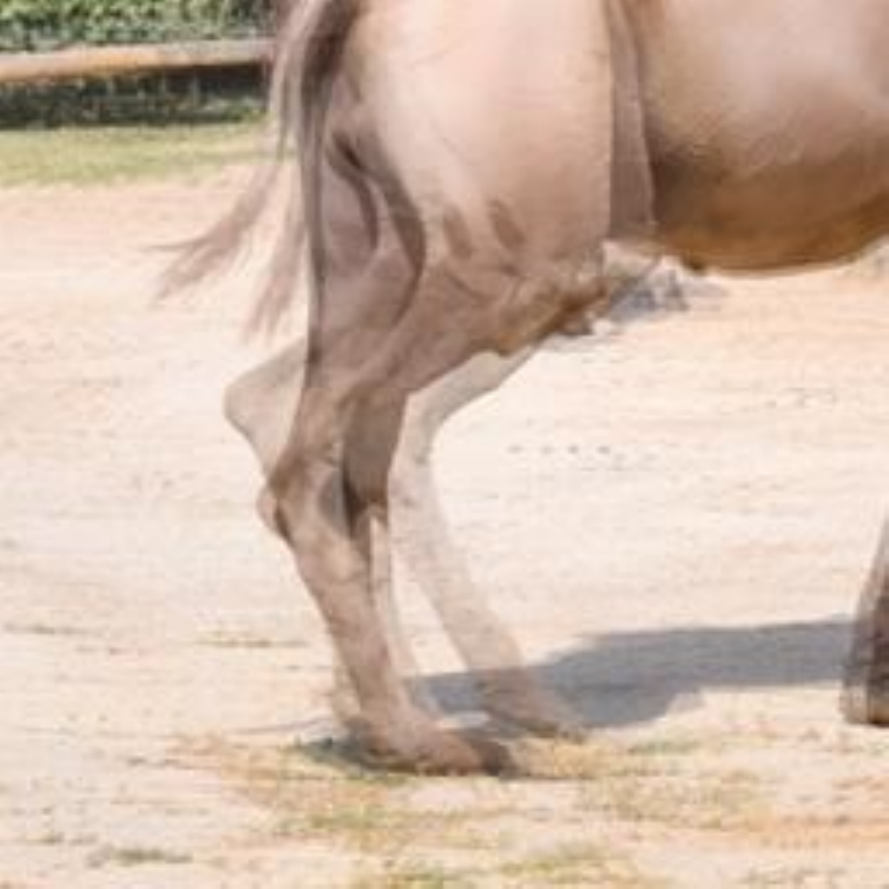}}\;\!\!
    \subfloat[$\bar{G}^{l=0}_{t\rightarrow 1}$] {\includegraphics[width=0.138\linewidth]{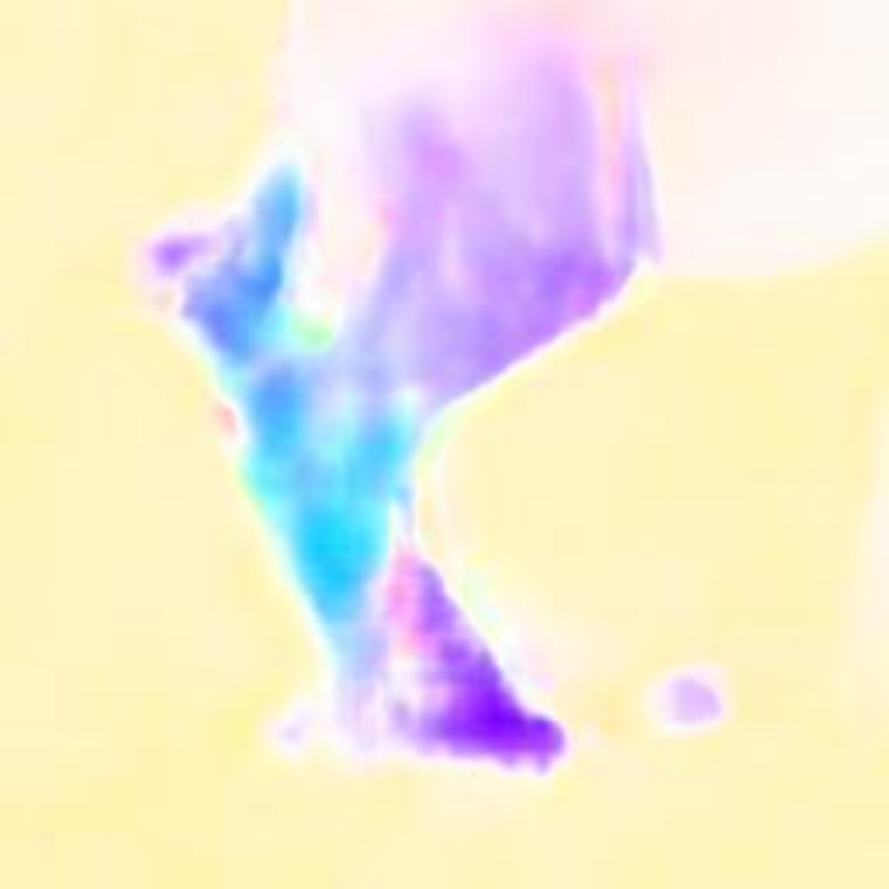}}\;\!\!
	\subfloat[$\bar{G}^{l=0}_{t\rightarrow 2}$] {\includegraphics[width=0.138\linewidth]{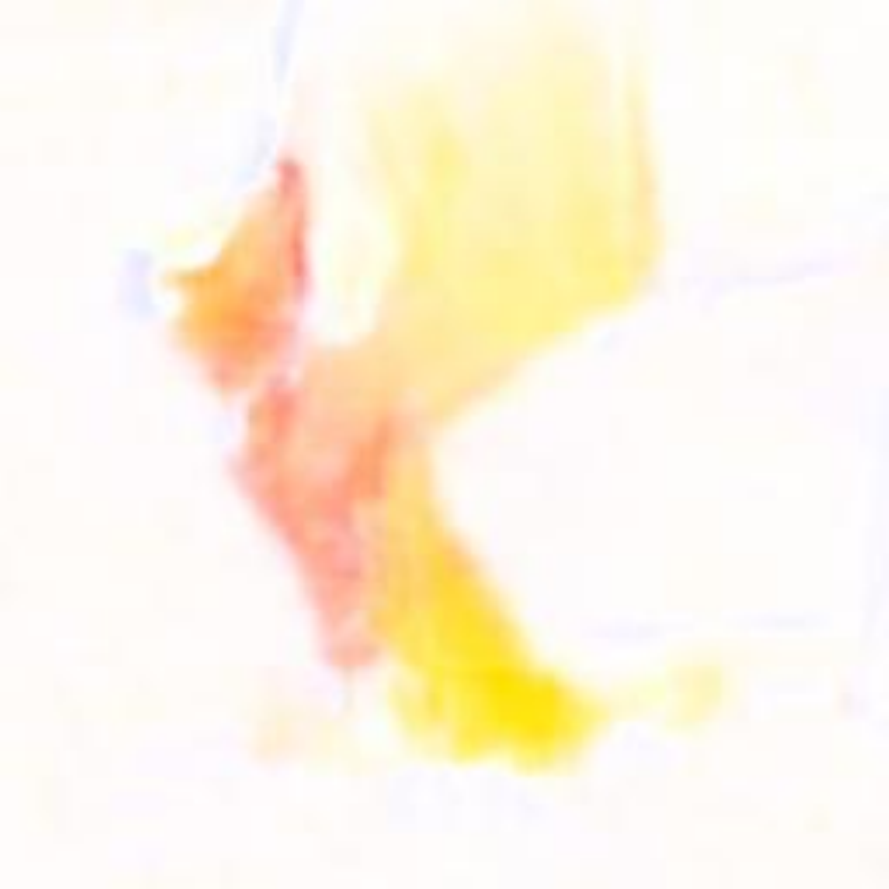}}\;\!\!
    \subfloat[$\bar{G}^{l=1}_{t\rightarrow 1}$] {\includegraphics[width=0.138\linewidth]{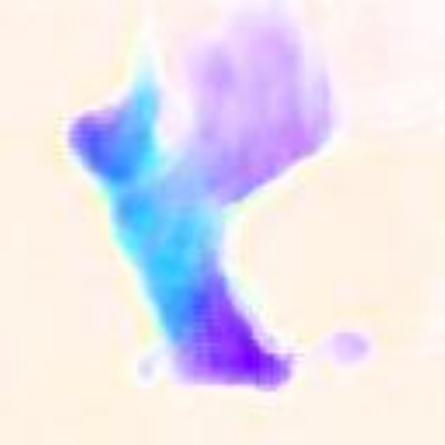}}\;\!\!
    \subfloat[$\bar{G}^{l=1}_{t\rightarrow 2}$] {\includegraphics[width=0.138\linewidth]{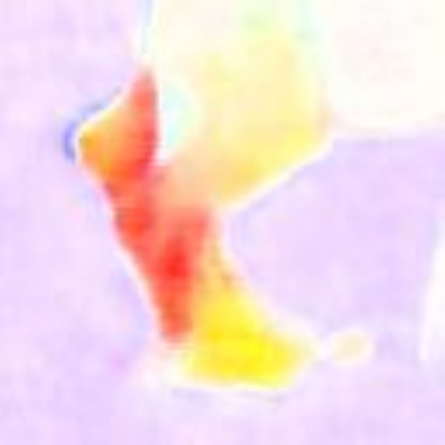}}\;\!\!
    \subfloat[$\bar{G}^{l=-1}_{t\rightarrow 1}$] {\includegraphics[width=0.138\linewidth]{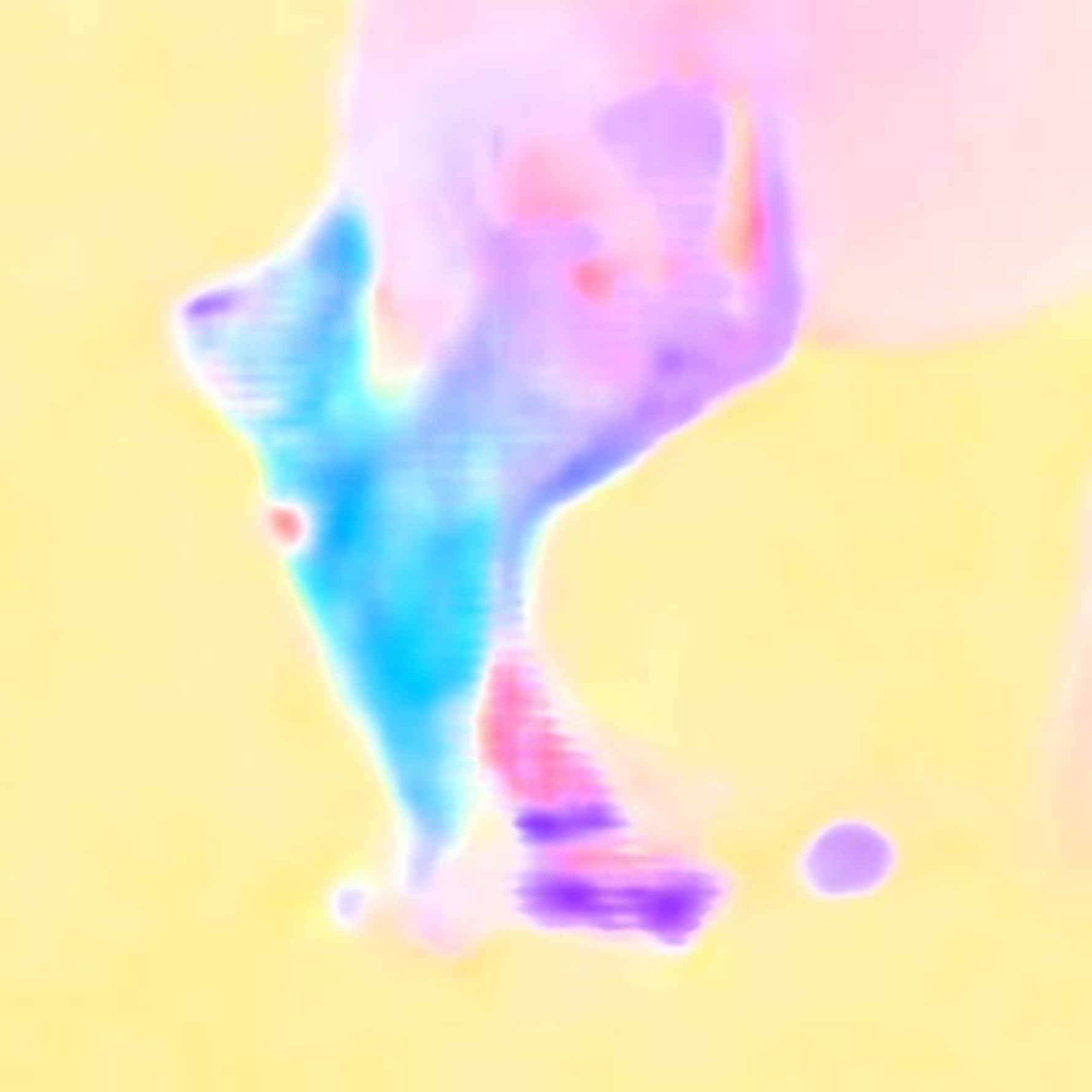}}\;\!\!
    \subfloat[$\bar{G}^{l=-1}_{t\rightarrow 2}$] {\includegraphics[width=0.138\linewidth]{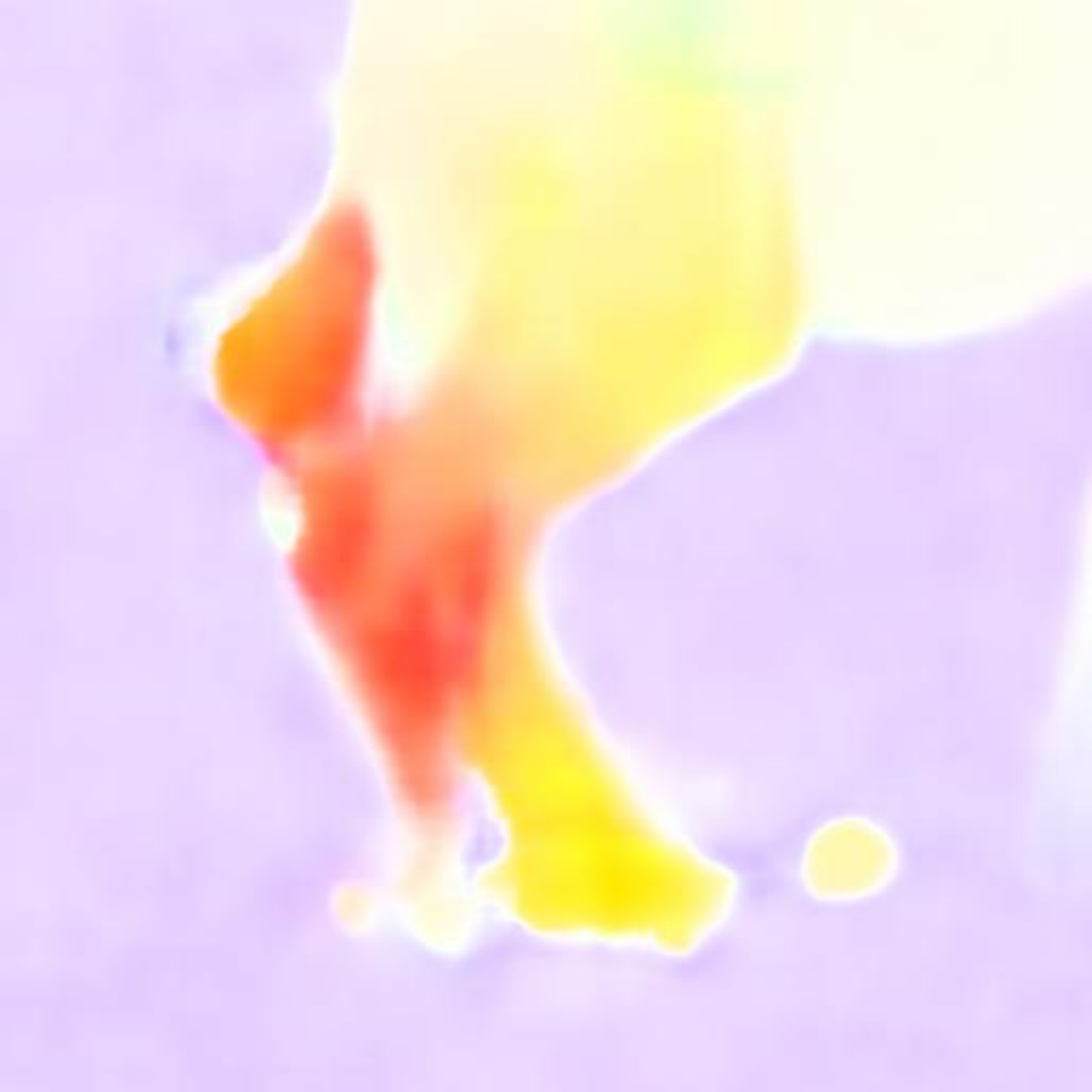}}
    \caption{\label{fig:motion} Visualization of the multi-scale multi-flows predicted by the network.}
\end{figure*}

\noindent\textbf{ST-GAN}. In the main paper, due to space limitations, we only evaluated the effectiveness of the adopted ST-GAN using visual examples. Here we additionally present the quantitative ablation study results for the adopted ST-GAN. For this purpose, we evaluate the same variants of ST-MFNet as described in Section 5.1 (the ST-GAN sub-section) of the main paper, that is, the distortion-oriented version (Our-$\mathcal{L}_{lap}$), the version fine-tuned with ST-GAN (Our-$\mathcal{L}_p$), the version fine-tuned with FIGAN~\cite{lee2020adacof} and the version fine-tuned with TGAN~\cite{saito2017temporal}. Table~\ref{tab:stgan} summarizes the performance of these variants on all four test sets in terms of LPIPS. It can be clearly observed from the table that the ST-GAN adopted in our work provides the best overall LPIPS performance, indicating its effectiveness for enhancing perceptual quality of the interpolated results.

\section{Visualization of Motion Fields}\label{sec:visualize}
To better understand the effectiveness of the multi-scale multi-flow estimation in the MIFNet branch, the predicted multi-flows are visualized here in the same manner as done in \cite{lee2020adacof}. That is, the mean flow maps at scale $l$, $\bar{G}^l_{t\rightarrow n}$ (where $n=1,2$), are obtained using Equations (\ref{eqn1}) and (\ref{eqn2}), and shown in Figure~\ref{fig:motion}. Note that for the purpose of visualization, the flows at the down- and up-sampled scales are re-scaled to the original resolution using the nearest neighbor filter.
\begin{gather}
    \mathbf{g}(x,y,i) = (\boldsymbol{\alpha}(x,y,i), \boldsymbol{\beta}(x,y,i)) \label{eqn1}\\
    \bar{G}^l_{t\rightarrow n}(x,y) = \sum_{i=1}^N \mathbf{w}(x,y,i) \mathbf{g}(x,y,i)\label{eqn2}
\end{gather}
It can be observed from Figure~\ref{fig:motion} that compared to the mean flow map at the original scale ($l=0$), the flows estimated for the down-sampled scale ($l=1$) tend to depict the general motion coarsely in different regions. On the other hand, the flow maps at the up-sampled scale ($l=-1$) reflect more detailed motion information.

\section{Comprehensive Evaluation Results}
\label{sec:lpips}

In the main paper, we presented our quantitative evaluation results of the proposed ST-MFNet and 14 competing methods in terms of PSNR and SSIM. Here, we additionally evaluate these methods in terms of LPIPS. The full results on the test sets UCF101~\cite{soomro2012ucf101}, DAVIS~\cite{perazzi2016benchmark} and VFITex are summarized in Table~\ref{tab:lpips1}, and the results on SNU-FILM~\cite{choi2020channel} are shown in Table~\ref{tab:lpips2}.

\section{Results of 4$\times$ and 8$\times$ Interpolation} \label{sec:4x8x}
The performance of the proposed ST-MFNet on multi-frame interpolation task is also evaluated, and compared to three best-performing benchmark algorithms: QVI~\cite{xu2019quadratic}, FLAVR~\cite{kalluri2020flavr} and Softsplat~\cite{niklaus2020softmax}. The algorithms were applied recursively to generate all the intermediate frames. The 11 test sequences at 240 FPS in the GoPro dataset~\cite{nah2017deep} were used as the test set for 4$\times$ and 8$\times$ interpolation. Table~\ref{tab:x4x8} summarizes the results, where it can be seen that ST-MFNet shows the best overall performance.

\section{Validation of Model Design} \label{sec:adacof+softsplat}
The proposed ST-MFNet combines multi-flow and single-flow based warping methods to enhance the interpolation quality of both complex and large motions. A natural question to ask is whether the performance of the model comes from the specific model design or simply from ensembling effect. To address this question, we create an ensemble model as a baseline, which simply combines AdaCoF~\cite{lee2020adacof} and Softsplat~\cite{niklaus2020softmax} through arithmetic averaging. This baseline model was trained under the same configurations as ST-MFNet and compared to the latter quantitatively. The results are summarized in Table~\ref{tab:ensemble}, where it is noted that although ensembling of AdaCoF and Softsplat does provide some benefit, the gain is marginal. This implies that the main source of the performance gain in ST-MFNet is the model design.

\section{User Study}
\label{sec:study}

The user study was conducted in a darkened, lab-based environment. The test sequences were played on a SONY PVM-X550 display, with screen size 124.2$\times$71.8cm. The display resolutions were configured to 1920$\times$1080 (spatial) and 60Hz (temporal), and the viewing distance was 2.15 meters (three times the screen height)~\cite{itu2002500}. The presentation of video sequences was controlled by a Windows PC running Matlab Psychtoolbox~\cite{pychotoolbox}. In each trial, a pair of videos to be compared were played twice, then the participant was asked to select the video with better perceived quality through an interface developed using the Psychtoolbox. This user study and the use of human data have undergone an internal ethics review and has been approved by the Institutional Review Board.

\section{Video Demo} \label{sec:video}
A video containing interpolation examples generated by ST-MFNet and more visual comparisons is available via this link: \url{https://drive.google.com/file/d/1zpE3rCQNJi4e8ADNWKbJA5wTvPllKZSj/view?usp=sharing}.

\section{Attribution of Assets}
\label{sec:license}

The data and code assets employed in this work and their corresponding license information are summarized in Table~\ref{tab:license1} and \ref{tab:license2} respectively.
\begin{table*}[t]
	\begin{center}

\resizebox{\linewidth}{!}{\begin{tabular}{lccccccccc}
\toprule
 & \multicolumn{3}{c}{UCF101} & \multicolumn{3}{c}{DAVIS}& \multicolumn{3}{c}{VFITex}\\ \cmidrule(lr){2-4}  \cmidrule(lr){5-7}  \cmidrule(lr){8-10}
 & PSNR ($\uparrow$) & SSIM ($\uparrow$) & LPIPS ($\downarrow$) & PSNR ($\uparrow$) & SSIM ($\uparrow$) & LPIPS ($\downarrow$) & PSNR ($\uparrow$) & SSIM ($\uparrow$) & LPIPS ($\downarrow$) \\ \midrule
DVF~\cite{liu2017video} & 32.251 & 0.965 & 0.036 & 20.403 & 0.673 & 0.274 & 19.946 & 0.709 & 0.389 \\
SuperSloMo~\cite{jiang2018super} & 32.547 & 0.968 & \textcolor{blue}{0.028} & 26.523 & 0.866 & 0.119 & 27.914 & 0.911 & 0.217 \\
SepConv~\cite{niklaus2017video} & 32.524 & 0.968 & 0.035 & 26.441 & 0.853 & 0.169 & 27.635 & 0.907 & 0.230 \\
DAIN~\cite{bao2019depth} & \underline{32.524} & \underline{0.968} & \underline{0.030} & 27.086 & 0.873 & \textcolor{blue}{0.117} & 27.314 & 0.909 & \textcolor{blue}{0.212} \\
BMBC~\cite{park2020bmbc} & \underline{32.729} & \underline{0.969} & \underline{0.032} & 26.835 & 0.869 & 0.125 & 27.337 & 0.904 & 0.220 \\
AdaCoF~\cite{lee2020adacof} & \underline{32.610} & \underline{0.968} & \underline{0.033} & 26.445 & 0.854 & 0.158 & 27.639 & 0.904 & 0.222 \\
FeFlow~\cite{gui2020featureflow} & 32.520 & 0.967 & 0.036 & 26.555 & 0.856 & 0.169 & OOM & OOM & OOM \\
CDFI~\cite{ding2021cdfi} & \underline{32.653} & \underline{0.968} & \underline{\textcolor{red}{0.024}} & 26.471 & 0.857 & 0.157 & 27.576 & 0.906 & 0.218 \\
CAIN~\cite{choi2020channel} & \underline{32.537} & \underline{0.968} & \underline{0.037} & 26.477 & 0.857 & 0.197 & 28.184 & 0.911 & 0.240 \\
Softsplat~\cite{niklaus2020softmax} & 32.835 & 0.969 & 0.037 & 27.582 & 0.881 & \textcolor{red}{0.116} & 28.813 & 0.924 & 0.221 \\
EDSC~\cite{cheng2021multiple} & \underline{32.677} & \underline{0.969} & \underline{0.033} & 26.968 & 0.860 & 0.142 & 27.641 & 0.904 & 0.222 \\
XVFI~\cite{sim2021xvfi} & 32.224 & 0.966 & 0.038 & 26.565 & 0.863 & 0.125 & 27.759 & 0.909 & 0.218 \\
QVI~\cite{xu2019quadratic} & 32.668 & 0.967 & 0.036 & 27.483 & 0.883 & 0.181 & 28.819 & \textcolor{blue}{0.926} & \textcolor{red}{0.210} \\
FLAVR~\cite{kalluri2020flavr} & \underline{\textcolor{red}{33.389}} & \underline{\textcolor{red}{0.971}} & \underline{0.035} & 27.450 & 0.873 & 0.190 & 28.487 & 0.915 & 0.233\\
\midrule
ST-MFNet (Ours-$\mathcal{L}_{lap}$) & \textcolor{blue}{33.384} & \textcolor{blue}{0.970} & 0.036 & \textcolor{red}{28.287} & \textcolor{red}{0.895} & 0.125 & \textcolor{red}{29.175} & \textcolor{red}{0.929} & 0.216 \\
ST-MFNet (Ours-$\mathcal{L}_p$) & 33.364 & \textcolor{blue}{0.970} & 0.033 & \textcolor{blue}{28.172} & \textcolor{blue}{0.892} & \textcolor{red}{0.116} & \textcolor{blue}{28.945} & 0.924 & \textcolor{red}{0.210} \\
\bottomrule
\end{tabular}}
\caption{\label{tab:lpips1}Quantitative comparison results for our model and 14 tested methods on UCF101, DAVIS and VFITex, in terms of PSNR, SSIM and LPIPS. OOM denotes cases where our GPU runs out of memory for the evaluation. For each column, the best result is colored in \textcolor{red}{red} and the second best is colored in \textcolor{blue}{blue}. Underlined scores denote the performance of pre-trained models rather than our re-trained versions.}
	\end{center}
\end{table*}

\begin{table*}[t]
	\begin{center}

\resizebox{\linewidth}{!}{\begin{tabular}{lcccccccccccc}
\toprule
 & \multicolumn{12}{c}{SNU-FILM} \\\cmidrule(l{5pt}r{5pt}){2-13}
 & \multicolumn{3}{c}{Easy} & \multicolumn{3}{c}{Medium} & \multicolumn{3}{c}{Hard} & \multicolumn{3}{c}{Extreme}\\ \cmidrule(lr){2-4}  \cmidrule(lr){5-7}  \cmidrule(lr){8-10} \cmidrule(lr){11-13}
 & PSNR  & SSIM  & LPIPS  & PSNR  & SSIM  & LPIPS  & PSNR  & SSIM  & LPIPS  & PSNR  & SSIM  & LPIPS  \\ \midrule
DVF~\cite{liu2017video} &  27.528 & 0.876 & 0.109 & 24.091 & 0.817 & 0.166 & 21.556 & 0.760 & 0.231 & 19.709 & 0.705 & 0.303  \\
SuperSloMo~\cite{jiang2018super} &  36.255 & 0.984 & 0.025 & 33.802 & 0.973 & 0.034 & 29.519 & 0.930 & 0.068 & 24.770 & 0.855 & 0.141 \\
SepConv~\cite{niklaus2017video} & 39.894 & 0.990 & 0.022 & 35.264 & 0.976 & 0.043 & 29.620 & 0.926 & 0.094 & 24.653 & 0.851 & 0.183\\
DAIN~\cite{bao2019depth} & 39.280 & 0.989 & 0.020 & 34.993 & 0.976 & 0.033 & 29.752 & 0.929 & 0.082 & 24.819 & 0.850 & 0.142\\
BMBC~\cite{park2020bmbc} & 39.809 & 0.990 & 0.020 & 35.437 & 0.978 & 0.034 & 29.942 & 0.933 & 0.088 & 24.715 & 0.856 & 0.145\\
AdaCoF~\cite{lee2020adacof} & 39.912 & 0.990 & 0.021 & 35.269 & 0.977 & 0.039 & 29.723 & 0.928 & 0.080 & 24.656 & 0.851 & 0.152\\
FeFlow~\cite{gui2020featureflow} & 39.591 & 0.990 & 0.022 & 35.014 & 0.977 & 0.041 & 29.466 & 0.928 & 0.090 & 24.607 & 0.852 & 0.182\\
CDFI~\cite{ding2021cdfi} & 39.881 & 0.990 & \textcolor{blue}{0.019} & 35.224 & 0.977 & 0.036 & 29.660 & 0.929 & 0.081 & 24.645 & 0.854 & 0.163\\
CAIN~\cite{choi2020channel} & \underline{39.890} & \underline{0.990} & \underline{0.021} & 35.630 & 0.978 & 0.037 & 29.998 & 0.931 & 0.097 & 25.060 & 0.857 & 0.203\\
Softsplat~\cite{niklaus2020softmax} & 40.165 & \textcolor{blue}{0.991} & 0.021 & 36.017 & 0.979 & 0.036 & 30.604 & 0.937 & \textcolor{blue}{0.066} & 25.436 & 0.864 & \textcolor{red}{0.119}\\
EDSC~\cite{cheng2021multiple} & 39.792 & 0.990 & 0.023 & 35.283 & 0.977 & 0.040 & 29.815 & 0.929 & 0.080 & 24.872 & 0.854 & 0.153\\
XVFI~\cite{sim2021xvfi} & 38.849 & 0.989 & 0.022 & 34.497 & 0.975 & 0.039 & 29.381 & 0.929 & 0.075 & 24.677 & 0.855 & \textcolor{blue}{0.139}\\
QVI~\cite{xu2019quadratic} & 36.648 & 0.985 & \textcolor{blue}{0.019} & 34.637 & 0.978 & \textcolor{blue}{0.032} & 30.614 & 0.947 & \textcolor{blue}{0.066} & 25.426 & 0.866 & 0.140\\
FLAVR~\cite{kalluri2020flavr} & 40.135 & 0.990 & 0.021 & 35.988 & 0.979 & 0.049 & 30.541 & 0.937 & 0.112 & 25.188 & 0.860 & 0.218\\
\midrule
ST-MFNet (Ours-$\mathcal{L}_{lap}$) & \textcolor{red}{40.775} & \textcolor{red}{0.992} & \textcolor{blue}{0.019} & \textcolor{red}{37.111} & \textcolor{red}{0.985} & 0.036 & \textcolor{red}{31.698} & \textcolor{red}{0.951} & 0.073 & \textcolor{red}{25.810} & \textcolor{red}{0.874} & 0.148\\
ST-MFNet (Ours-$\mathcal{L}_p$) & \textcolor{blue}{40.542} & \textcolor{blue}{0.991} & \textcolor{red}{0.017} & \textcolor{blue}{36.964} & \textcolor{blue}{0.983} & \textcolor{red}{0.031} & \textcolor{blue}{31.580} & \textcolor{blue}{0.949} & \textcolor{red}{0.065} & \textcolor{blue}{25.764} & \textcolor{blue}{0.871} & 0.140\\
\bottomrule
\end{tabular}}
\caption{\label{tab:lpips2}Quantitative comparison results for our model and 14 tested methods on SNU-FILM dataset, in terms of PSNR, SSIM and LPIPS. For each column, the best result is colored in \textcolor{red}{red} and the second best is colored in \textcolor{blue}{blue}. Underlined scores denote the performance of pre-trained models rather than our re-trained versions.}
	\end{center}
\end{table*}

\begin{table*}[t]
\begin{center}

\begin{tabular}{lcccccc}
\toprule
 & \multicolumn{3}{c}{GoPro-4$\times$} & \multicolumn{3}{c}{GoPro-8$\times$} \\
 \cmidrule(lr){2-4}  \cmidrule(lr){5-7}  
 & PSNR ($\uparrow$) & SSIM ($\uparrow$) & LPIPS ($\downarrow$) & PSNR ($\uparrow$) & SSIM ($\uparrow$) & LPIPS ($\downarrow$)  \\ \midrule
QVI~\cite{xu2019quadratic} & \textcolor{blue}{29.324} & \textcolor{blue}{0.927} & \textcolor{red}{0.049} & 29.280 & \textcolor{blue}{0.929} & \textcolor{red}{0.048}\\
FLAVR~\cite{kalluri2020flavr} & 28.911 & 0.914 & 0.110 & 29.512 & 0.922 & 0.101\\
Softsplat~\cite{niklaus2020softmax} & 28.858 & 0.908 & \textcolor{blue}{0.072} & \textcolor{blue}{29.663} & 0.918 & \textcolor{blue}{0.067}\\
\midrule
ST-MFNet (Ours-$\mathcal{L}_{lap}$) & \textcolor{red}{29.892} & \textcolor{red}{0.926} & 0.098 & \textcolor{red}{30.568} & \textcolor{red}{0.934} & 0.092 \\
\bottomrule
\end{tabular}
\caption{\label{tab:x4x8}Quantitative comparison results for 4$\times$ and 8$\times$ interpolation on GoPro dataset in terms of PSNR, SSIM and LPIPS. For each column, the best result is colored in \textcolor{red}{red} and the second best is colored in \textcolor{blue}{blue}.}
\end{center}
\end{table*}

\begin{table*}[t]
\begin{center}
\begin{tabular}{l|c|r|cccc}
\toprule
\multicolumn{2}{l|}{}     &                    & AdaCoF & Softsplat  & AdaCoF+Softsplat & ST-MFNet (ours-$\mathcal{L}_{lap}$) \\ \midrule          
\multicolumn{2}{c|}{\multirow{2}{*}{UCF101}} & PSNR ($\uparrow$) & 32.488& 32.683& \textcolor{blue}{32.729}&\textcolor{red}{33.384}\\
\multicolumn{2}{l|}{}                        & SSIM ($\uparrow$) & 0.968& \textcolor{blue}{0.969}& \textcolor{blue}{0.969}&\textcolor{red}{0.970}\\
\midrule
\multicolumn{2}{c|}{\multirow{2}{*}{DAVIS}} & PSNR ($\uparrow$) & 26.445& 27.359& \textcolor{blue}{27.361}&\textcolor{red}{28.287}\\
\multicolumn{2}{l|}{}                       & SSIM ($\uparrow$) & 0.854& \textcolor{blue}{0.878}& \textcolor{blue}{0.878}&\textcolor{red}{0.895}\\
\midrule                  
\multirow{8}{*}{SNU-FILM}&\multirow{2}{*}{Easy}   & PSNR ($\uparrow$)& 39.912& 40.021& \textcolor{blue}{40.083} &\textcolor{red}{40.775}\\
                          &                        & SSIM ($\uparrow$)& 0.990& \textcolor{blue}{0.991}& \textcolor{blue}{0.991} &\textcolor{red}{0.992}\\
                        \cmidrule{2-7}
                          &\multirow{2}{*}{Medium} & PSNR  ($\uparrow$)& 35.269& 35.833& \textcolor{blue}{35.841} &\textcolor{red}{37.111}\\
                          &                        & SSIM ($\uparrow$)& 0.977& \textcolor{blue}{0.979}& \textcolor{blue}{0.979} &\textcolor{red}{0.985}\\ 
                          \cmidrule{2-7}
                          &\multirow{2}{*}{Hard}   & PSNR ($\uparrow$) & 29.723& 30.412& \textcolor{blue}{30.449} &\textcolor{red}{31.698}\\
                          &                        & SSIM ($\uparrow$)& 0.928& 0.936& \textcolor{blue}{0.937} &\textcolor{red}{0.951}\\
                        \cmidrule{2-7}
                          &\multirow{2}{*}{Extreme} & PSNR ($\uparrow$) & 24.656& 25.242&\textcolor{blue}{25.258} &\textcolor{red}{25.810}\\
                          &                        & SSIM ($\uparrow$)& 0.851& 0.862& \textcolor{blue}{0.864} &\textcolor{red}{0.874}\\
                        \midrule
\multicolumn{2}{c|}{\multirow{2}{*}{VFITex}} & PSNR ($\uparrow$) & 27.639& 28.620& \textcolor{blue}{28.629}&\textcolor{red}{29.175}\\
\multicolumn{2}{l|}{}                        & SSIM ($\uparrow$) & 0.904& 0.922& \textcolor{blue}{0.923}&\textcolor{red}{0.929}\\
\bottomrule
\end{tabular}
\caption{\label{tab:ensemble}Quantitative evaluation results of the proposed ST-MFNet and a simple baseline that combines AdaCoF and Softsplat. For each row, the best result is colored in \textcolor{red}{red} and the second best is colored in \textcolor{blue}{blue}. Note Softsplat here is trained with Charbonnier loss so that AdaCoF, Softsplat and the baseline only differ in model design.}
\end{center}
\vspace{-5mm}
\end{table*}

\begin{table*}[t]
\centering
\resizebox{\linewidth}{!}{\small\begin{tabular}{m{3cm}|m{9cm}|m{4cm}}
\toprule
\textbf{Dataset}  & \textbf{Dataset URL} &\textbf{License / Terms of Use} \\ 
\midrule
Vimeo-90k~\cite{xue2019video} & \url{http://toflow.csail.mit.edu} &MIT license. \\
\midrule
BVI-DVC~\cite{ma2020bvi} & \url{https://fan-aaron-zhang.github.io/BVI-DVC/} & All sequences are allowed for academic research.\\
\midrule
UCF101~\cite{soomro2012ucf101} & \url{https://www.crcv.ucf.edu/research/data-sets/ucf101/} & No explicit license terms, but compiled and made available for research use by the University of Central Florida. \\
\midrule
DAVIS~\cite{perazzi2016benchmark} & \url{https://davischallenge.org} & BSD license. \\
\midrule
SNU-FILM~\cite{choi2020channel} & \url{https://myungsub.github.io/CAIN/} & MIT license .\\
\midrule
Xiph~\cite{montgomery3xiph} & \url{https://media. xiph. org/video/derf} & Sequences used are available for research use. \\
\midrule
Mitch Martinez Free 4K Stock Footage~\cite{mitch} & \url{http:// mitchmartinez.com/free-4k-red-epic-stock-footage/} & Sequences used are available for research use. \\
\midrule
UVG database~\cite{mercat2020uvg} & \url{http://ultravideo.fi} & Non-commercial Creative Commons BY-NC license.\\
\midrule
Pexels~\cite{pexels} & \url{https://www.pexels.com/videos/} & All sequences are available for research use. \\
\bottomrule
\end{tabular}}
\caption{License information for the datasets used in this work.}
\label{tab:license1}
\end{table*}

\begin{table*}[t]
\centering
\resizebox{\linewidth}{!}{\small\begin{tabular}{l|p{9cm}|p{4cm}}
\toprule
\textbf{Method}  & \textbf{Source code URL} & \textbf{License / Teams of Use} \\ 
\midrule
DVF~\cite{liu2017video} & \url{https://github.com/liuziwei7/voxel-flow} & Non-commercial research and education only. \\
\midrule
SuperSloMo~\cite{jiang2018super} & \url{https://github.com/avinashpaliwal/Super-SloMo} & MIT license. \\
\midrule
SepConv~\cite{niklaus2017video} & \url{https://github.com/sniklaus/sepconv-slomo} & Academic purposes only. \\
\midrule
DAIN~\cite{bao2019depth} & \url{https://github.com/baowenbo/DAIN} & MIT license. \\
\midrule
BMBC~\cite{park2020bmbc} & \url{https://github.com/JunHeum/BMBC} & MIT license. \\
\midrule
AdaCoF~\cite{lee2020adacof} & \url{https://github.com/HyeongminLEE/AdaCoF-pytorch} & MIT license. \\
\midrule
FeFlow~\cite{gui2020featureflow} & \url{https://github.com/CM-BF/FeatureFlow} & MIT license. \\
\midrule
CAIN~\cite{choi2020channel} & \url{https://github.com/myungsub/CAIN} & MIT license. \\
\midrule
SoftSplat~\cite{niklaus2020softmax} & \url{https://github.com/sniklaus/softmax-splatting} & Academic purposes only. \\
\midrule
XVFI~\cite{sim2021xvfi} & \url{https://github.com/JihyongOh/XVFI} & Research and education only. \\
\midrule
FLAVR~\cite{kalluri2020flavr} & \url{https://github.com/tarun005/FLAVR} & Apache-2.0 License. \\
\bottomrule

\end{tabular}}
\caption{License information for the code assets used in this work.}
\label{tab:license2}
\end{table*}

\clearpage
\clearpage
{\small
\bibliographystyle{ieee_fullname}
\bibliography{egbib}
}

\end{document}